\documentclass[11pt,a4paper]{article}

\usepackage{times,latexsym}
\usepackage{url}
\usepackage[T1]{fontenc}
\usepackage[utf8]{inputenc}
\usepackage[acceptedWithA]{tacl2021v1}
\usepackage{microtype}
\usepackage{inconsolata}
\usepackage{graphicx}
\usepackage[most]{tcolorbox}
\usepackage{adjustbox}
\usepackage{chessfss}
\usepackage{booktabs}
\usepackage{amsfonts}
\usepackage{amsmath}
\usepackage{amssymb}
\usepackage{nicefrac}
\usepackage{colortbl}
\usepackage{subcaption}
\usepackage{xspace}
\IfFileExists{acro.sty}{\usepackage{acro}}{}
\usepackage{tabularx}
\usepackage{multirow}
\usepackage{enumitem}
\usepackage{placeins}
\usepackage{algorithm}
\usepackage{algpseudocode}
\usepackage[capitalise,noabbrev]{cleveref}
\usepackage{xparse}
\usepackage{tikz}
\usepackage{pgfplots}
\usepgfplotslibrary{fillbetween,groupplots}
\usetikzlibrary{calc,shapes.geometric}
\pgfplotsset{compat=1.18}
\newcommand{\circledtextset}[1]{}
\NewDocumentCommand{\circledtext}{s O{} m}{%
  \tikz[baseline=(char.base)]{
    \node[
      draw,
      fill=black,
      text=white,
      draw=black,
      circle,
      inner sep=0.18ex,
      line width=0.4pt,
      minimum size=1.55ex
    ] (char) {\fontsize{5.4}{5.4}\selectfont #3};
  }%
}
\usepackage{fontawesome}

\definecolor{cardink}{HTML}{111111}
\definecolor{cardmuted}{HTML}{6B6B6B}
\definecolor{cardp1}{HTML}{5879A3}   %
\definecolor{cardp2}{HTML}{111111}   %
\definecolor{cardboard}{HTML}{EFECE4}
\definecolor{cardpaper}{HTML}{FBFAF7}

\definecolor{cardbeige}{HTML}{F4EFE4}   %
\definecolor{cardink}{HTML}{111111}

\newcommand{\gcrule}[1]{\noindent{\color{cardink}\rule{\linewidth}{#1}}\par}
\newcommand{\gamecardboardwidth}{0.25\linewidth}
\newcommand{\gamecardruleswidth}{0.73\linewidth}
\newcommand{\gcmeta}[2]{%
  \begin{tabular}[t]{@{}r@{}}
    {\textsc{\scriptsize Action}}\quad{\ttfamily\small #1}\\[1pt]
    {\textsc{\scriptsize Goal}}\quad{\ttfamily\small #2}
  \end{tabular}%
}
\newcommand{\gcsetupicon}{%
  \raisebox{-0.35ex}{\begin{tikzpicture}[x=0.92em,y=0.92em,baseline=-0.35ex]
    \draw[cardmuted,line width=0.45pt] (0,0) rectangle (1,1);
    \draw[cardmuted,line width=0.35pt] (0.33,0)--(0.33,1) (0.66,0)--(0.66,1) (0,0.33)--(1,0.33) (0,0.66)--(1,0.66);
    \fill[cardmuted] (0.50,0.50) circle (0.09);
  \end{tikzpicture}}%
}
\newcommand{\gcmoveicon}{%
  \raisebox{-0.35ex}{\begin{tikzpicture}[x=0.92em,y=0.92em,baseline=-0.35ex]
    \draw[cardmuted,line width=0.65pt,-latex,rounded corners=1pt] (0.10,0.20)--(0.45,0.20)--(0.45,0.78)--(0.88,0.78);
  \end{tikzpicture}}%
}
\newcommand{\gcruleicon}{%
  \raisebox{-0.35ex}{\begin{tikzpicture}[x=0.92em,y=0.92em,baseline=-0.35ex]
    \draw[cardmuted,line width=0.45pt] (0.15,0.25)--(0.85,0.25) (0.15,0.50)--(0.85,0.50) (0.15,0.75)--(0.85,0.75);
    \fill[cardmuted] (0.38,0.25) circle (0.08) (0.62,0.50) circle (0.08) (0.46,0.75) circle (0.08);
  \end{tikzpicture}}%
}
\newcommand{\gcwinicon}{%
  \raisebox{-0.35ex}{\begin{tikzpicture}[x=0.92em,y=0.92em,baseline=-0.35ex]
    \draw[cardmuted,line width=0.55pt] (0.22,0.08)--(0.22,0.92);
    \filldraw[fill=cardmuted!18,draw=cardmuted,line width=0.45pt] (0.22,0.86)--(0.82,0.76)--(0.22,0.58)--cycle;
    \draw[cardmuted,line width=0.45pt] (0.10,0.08)--(0.48,0.08);
  \end{tikzpicture}}%
}

\newenvironment{gamerules}{%
  \begin{description}[leftmargin=2.15em,labelwidth=1.25em,labelsep=0.55em,itemsep=3pt,topsep=0pt,font=\normalfont]%
}{%
  \end{description}%
}
\newcommand{\gcsetup}{\item[\gcsetupicon]}
\newcommand{\gcmove}{\item[\gcmoveicon]}
\newcommand{\gcrulekey}{\item[\gcruleicon]}
\newcommand{\gcwin}{\item[\gcwinicon]}

\newenvironment{gamecard}[3]{%
  \begin{tcolorbox}[
    enhanced, sharp corners, boxrule=0.6pt,
    colback=cardbeige, colframe=cardink,
    left=14pt, right=14pt, top=10pt, bottom=10pt
  ]%
  \noindent
  \begin{minipage}[t]{0.58\linewidth}\vspace{0pt}%
    {\large\bfseries #1}\par
    \vspace{1pt}{\itshape\color{cardmuted}#2}%
  \end{minipage}\hfill
  \begin{minipage}[t]{0.38\linewidth}\vspace{0pt}\raggedleft
    {\color{cardmuted}#3}%
  \end{minipage}\par
  \vspace{4pt}\gcrule{0.4pt}\vspace{10pt}%
}{%
  \end{tcolorbox}%
}

\newcommand{\connectfourboard}[1][0.62]{%
  \begin{tikzpicture}[scale=#1]
    \fill[cardboard] (0,0) rectangle (7,6);
    \foreach \c in {0,...,6}{%
      \foreach \r in {0,...,5}{%
        \draw[fill=cardpaper, draw=black!18, line width=0.2pt]
              (\c+0.5,\r+0.5) circle (0.36);}}%
    \foreach \c/\r/\p in {%
      2/0/cardp2, 3/0/cardp1, 4/0/cardp1, 5/0/cardp2,
      2/1/cardp1, 3/1/cardp2, 4/1/cardp1,
      3/2/cardp1, 4/2/cardp2}{%
      \fill[\p] (\c+0.5,\r+0.5) circle (0.36);}%
    \draw[cardink, line width=0.4pt] (0,0) rectangle (7,6);
  \end{tikzpicture}%
}

\definecolor{qwenThreeLine}{HTML}{3B6FB6}
\definecolor{qwenThreeFiveLine}{HTML}{D69F23}
\definecolor{llamaLine}{HTML}{2F9A8F}
\definecolor{gemmaLine}{HTML}{6FAE3F}
\definecolor{geminiLine}{HTML}{C65F2E}
\definecolor{gptLine}{HTML}{C84D6B}
\definecolor{claudeLine}{HTML}{6D5A3B}
\definecolor{rqblue}{HTML}{ECF4FF}
\definecolor{rqgreen}{HTML}{EAF8EA}
\definecolor{rqyellow}{HTML}{FFF6C7}
\definecolor{rqplotbg}{HTML}{FAF7EF}
\definecolor{rqplotline}{HTML}{27384D}
\definecolor{tracePolicy}{HTML}{F03BD8}
\definecolor{traceValue}{HTML}{2FA7FF}
\definecolor{traceState}{HTML}{20B72A}
\definecolor{spiralLine}{HTML}{C65F2E}
\definecolor{baseLine}{HTML}{4F5661}
\definecolor{resultGain}{HTML}{167C3B}
\definecolor{resultDrop}{HTML}{B8323C}
\definecolor{qwenVersion}{HTML}{6D28D9}
\definecolor{tableDeltaBg}{HTML}{F2F2F2}
\definecolor{mypink}{RGB}{255, 105, 180}
\definecolor{myblue}{RGB}{0, 114, 189}
\definecolor{myyellow}{RGB}{255, 215, 0}
\definecolor{mypurple}{RGB}{155,89,182}
\definecolor{rqModelLlama}{rgb}{0.003922,0.450980,0.698039}
\definecolor{rqModelOcto}{rgb}{0.870588,0.560784,0.019608}
\definecolor{rqModelQwenFour}{rgb}{0.007843,0.619608,0.450980}
\definecolor{rqModelQwenEight}{rgb}{0.835294,0.368627,0.000000}

\newcommand{\qwenicon}{\raisebox{-0.35em}{\includegraphics[height=1.38em]{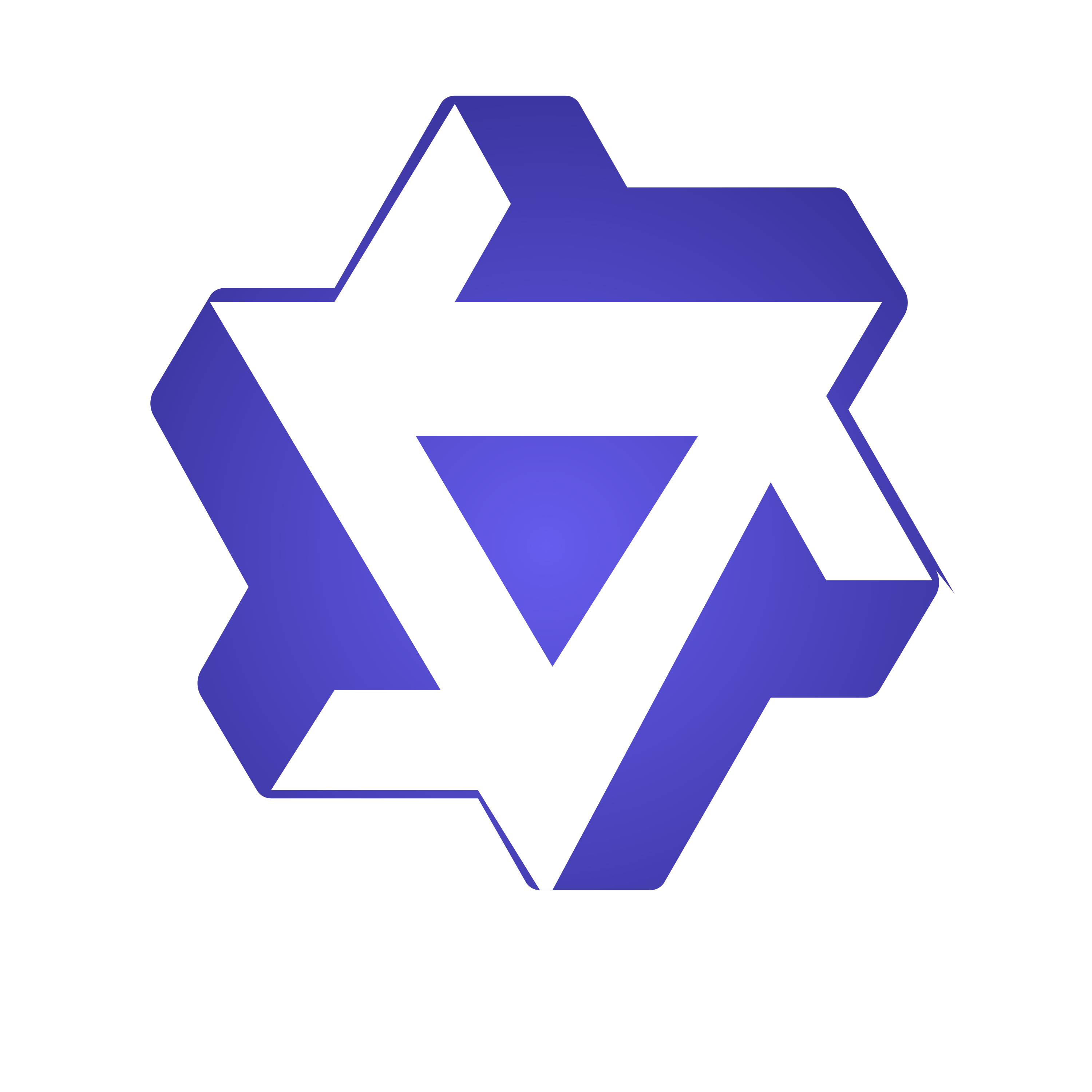}}\xspace}
\newcommand{\llamaicon}{\raisebox{-0.09em}{\includegraphics[height=0.94em]{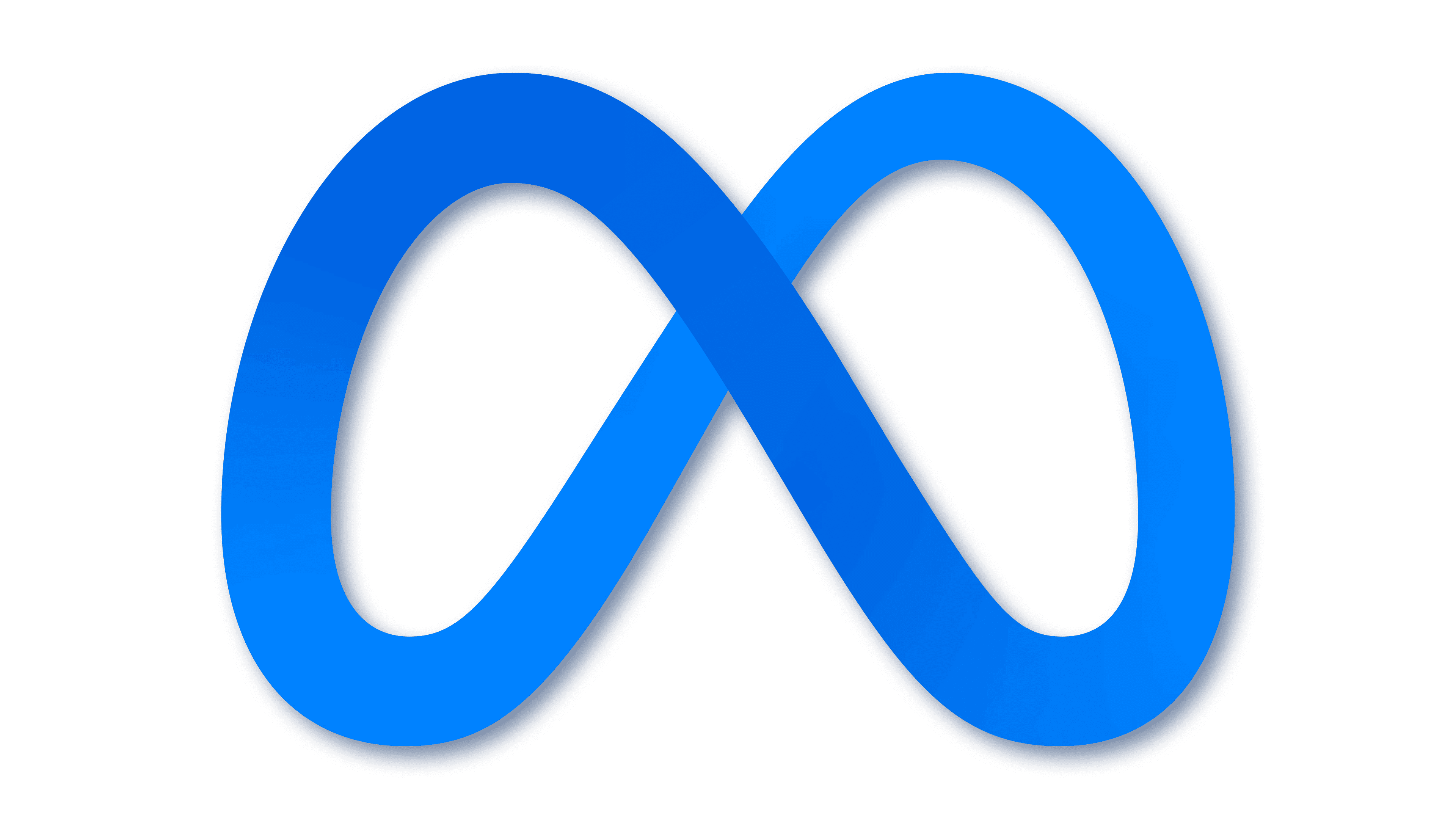}}\xspace}
\newcommand{\gemmaicon}{\raisebox{-0.18em}{\includegraphics[height=1.12em]{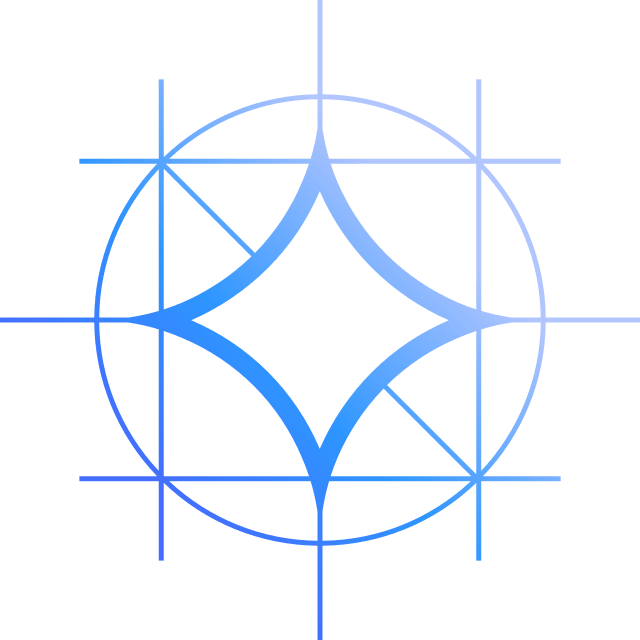}}\xspace}
\newcommand{\geminiicon}{\raisebox{-0.18em}{\includegraphics[height=1.12em]{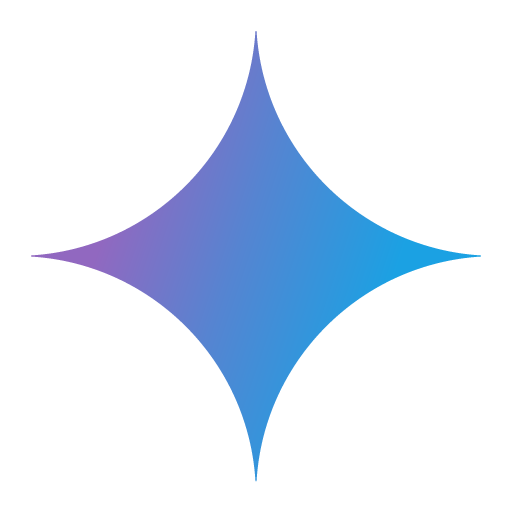}}\xspace}
\newcommand{\openaiicon}{\raisebox{-0.10em}{\includegraphics[height=1.06em]{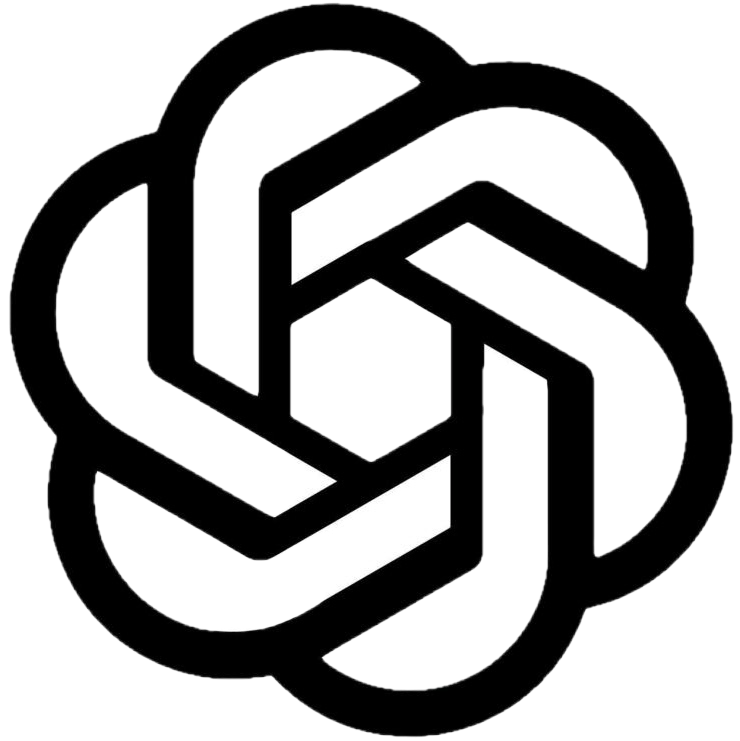}}\xspace}
\newcommand{\claudeicon}{\raisebox{-0.12em}{\includegraphics[height=1.05em]{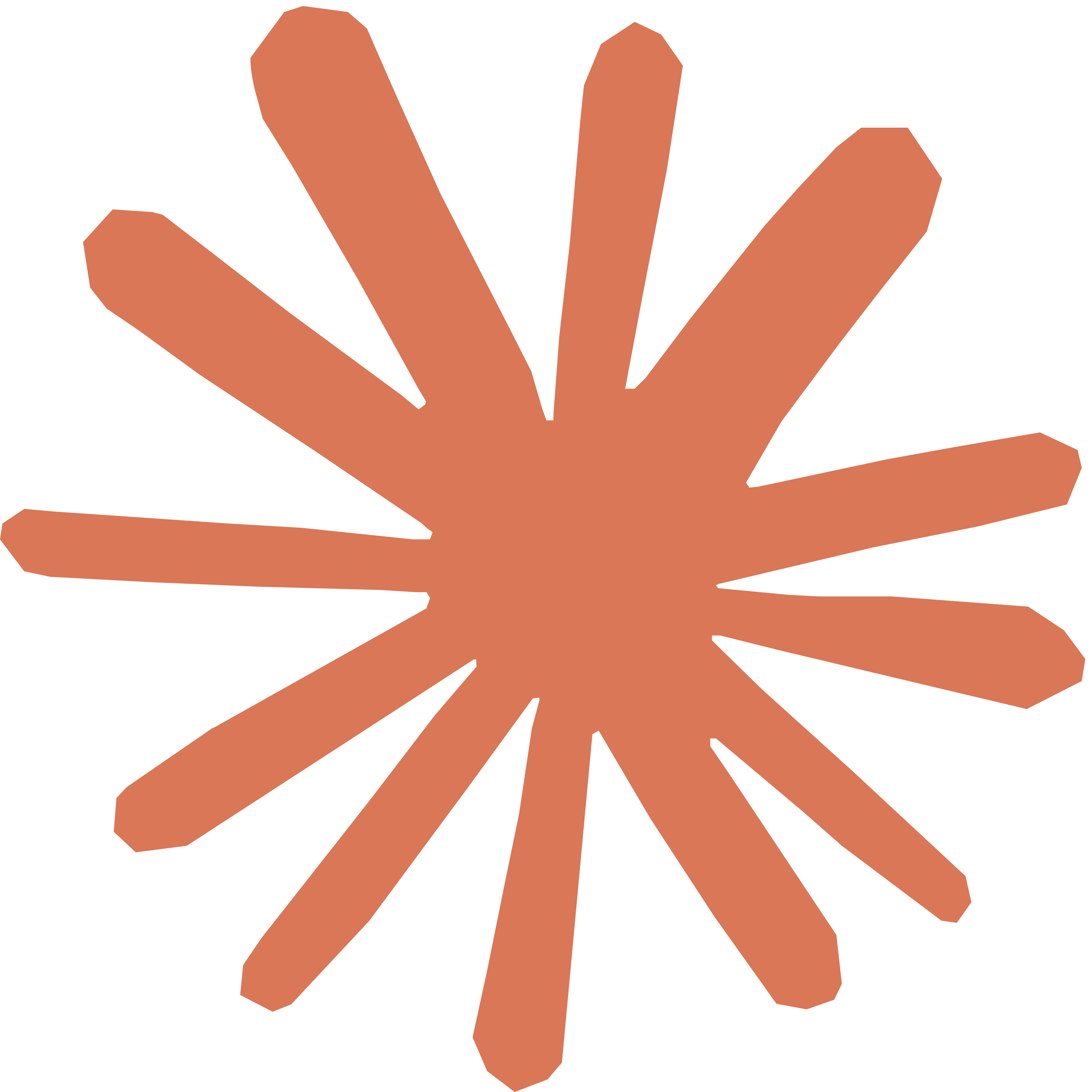}}\xspace}
\newcommand{\qwenploticon}{\includegraphics[width=0.66em]{qwen-icon.png}}
\newcommand{\llamaploticon}{\includegraphics[width=0.64em]{llama-icon.png}}
\newcommand{\gemmaploticon}{\includegraphics[width=0.64em]{gemma-icon.png}}
\newcommand{\geminiploticon}{\includegraphics[width=0.64em]{gemini-icon.png}}
\newcommand{\openaiploticon}{\includegraphics[width=0.63em]{openai-icon.png}}
\newcommand{\claudeploticon}{\includegraphics[width=0.64em]{claude-icon.png}}

\newcommand{\condtag}[3]{\begingroup\setlength{\fboxsep}{1.4pt}\colorbox{#2!14}{\textcolor{#2}{#1}\,\textsf{#3}}\endgroup}

\newcommand{\resultdelta}[2]{%
  \ensuremath{#1_{\scriptscriptstyle%
    \ifdim#2pt>0pt\textcolor{resultGain}{#2}%
    \else\ifdim#2pt<0pt\textcolor{resultDrop}{#2}%
    \else\textcolor{baseLine}{#2}%
    \fi\fi}}}

\newcommand{\best}[1]{{\bfseries\boldmath #1}}
\newcommand{\secondbest}[1]{\underline{#1}}
\newcommand{\basecond}{\condtag{$\varnothing$}{baseLine}{Baseline}}
\newcommand{\sftcond}{\condtag{$\blacktriangleright$}{tracePolicy}{SFT}}
\newcommand{\rulebotcond}{\condtag{$\blacktriangle$}{spiralLine}{RuleBot}}

\newcommand{\opsdcond}{\condtag{$\blacklozenge$}{traceState}{OPSD}}

\newcommand{\rqtag}[1]{\texttt{RQ-}\raisebox{-0.1em}{\includegraphics[height=0.75em]{#1}}\xspace}
\newcommand{\rqone}{\rqtag{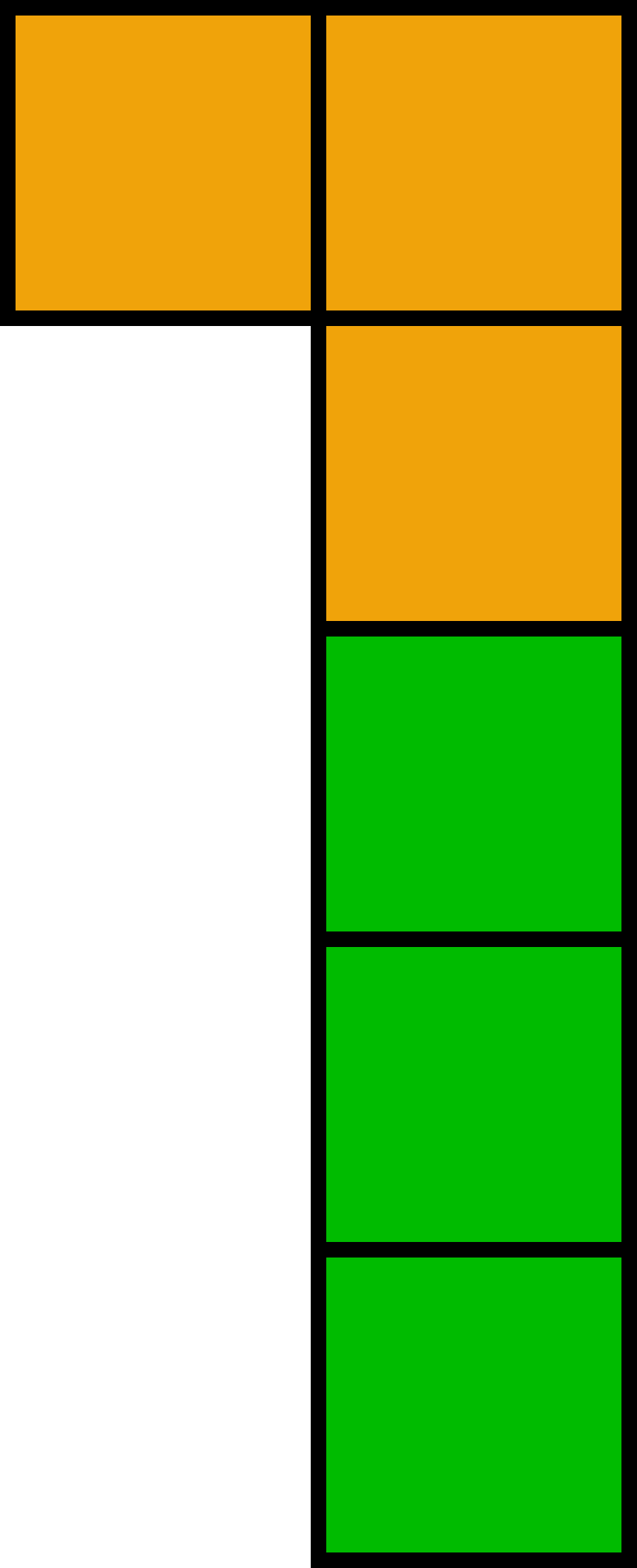}}
\newcommand{\rqtwo}{\rqtag{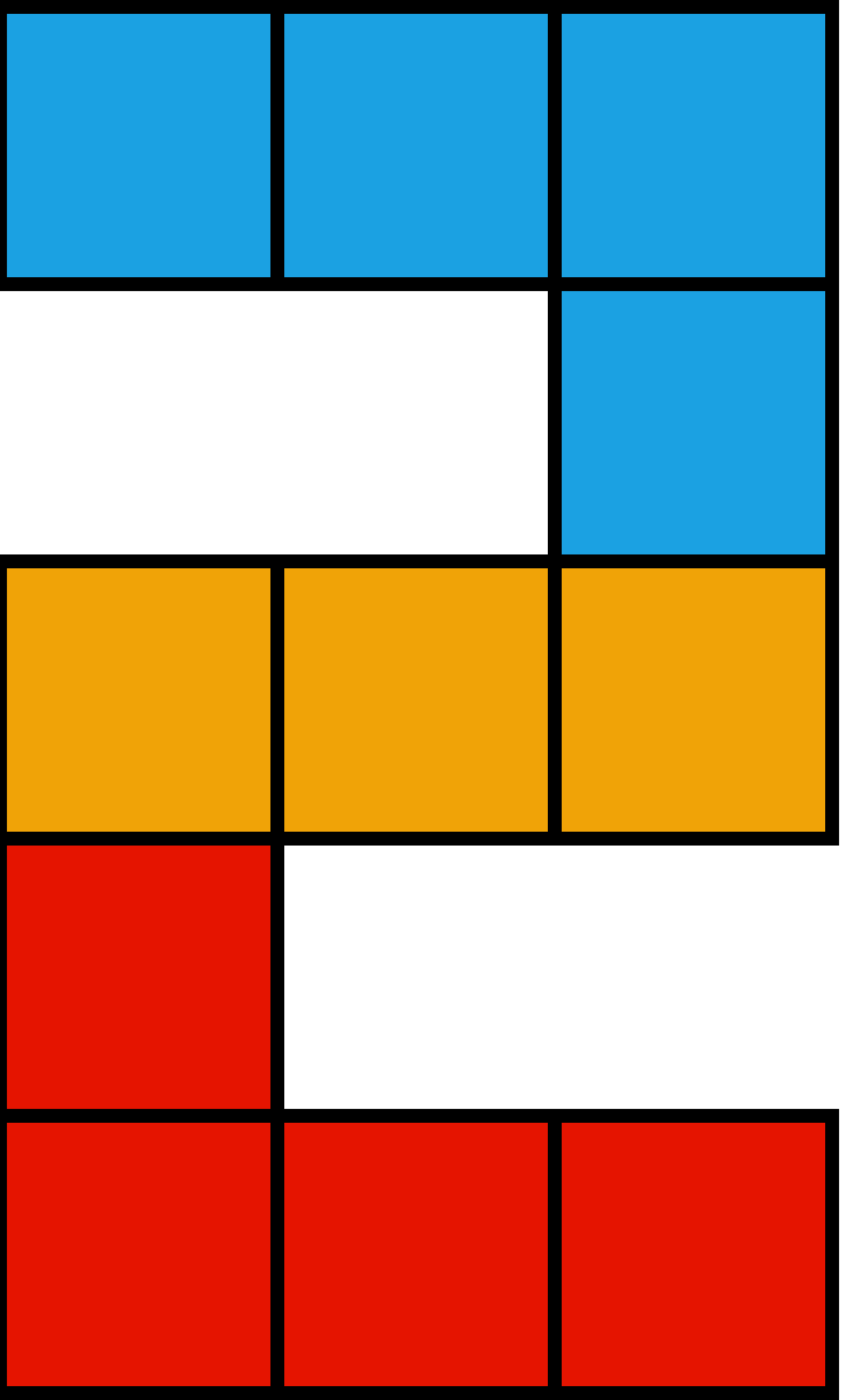}}
\newcommand{\rqthree}{\rqtag{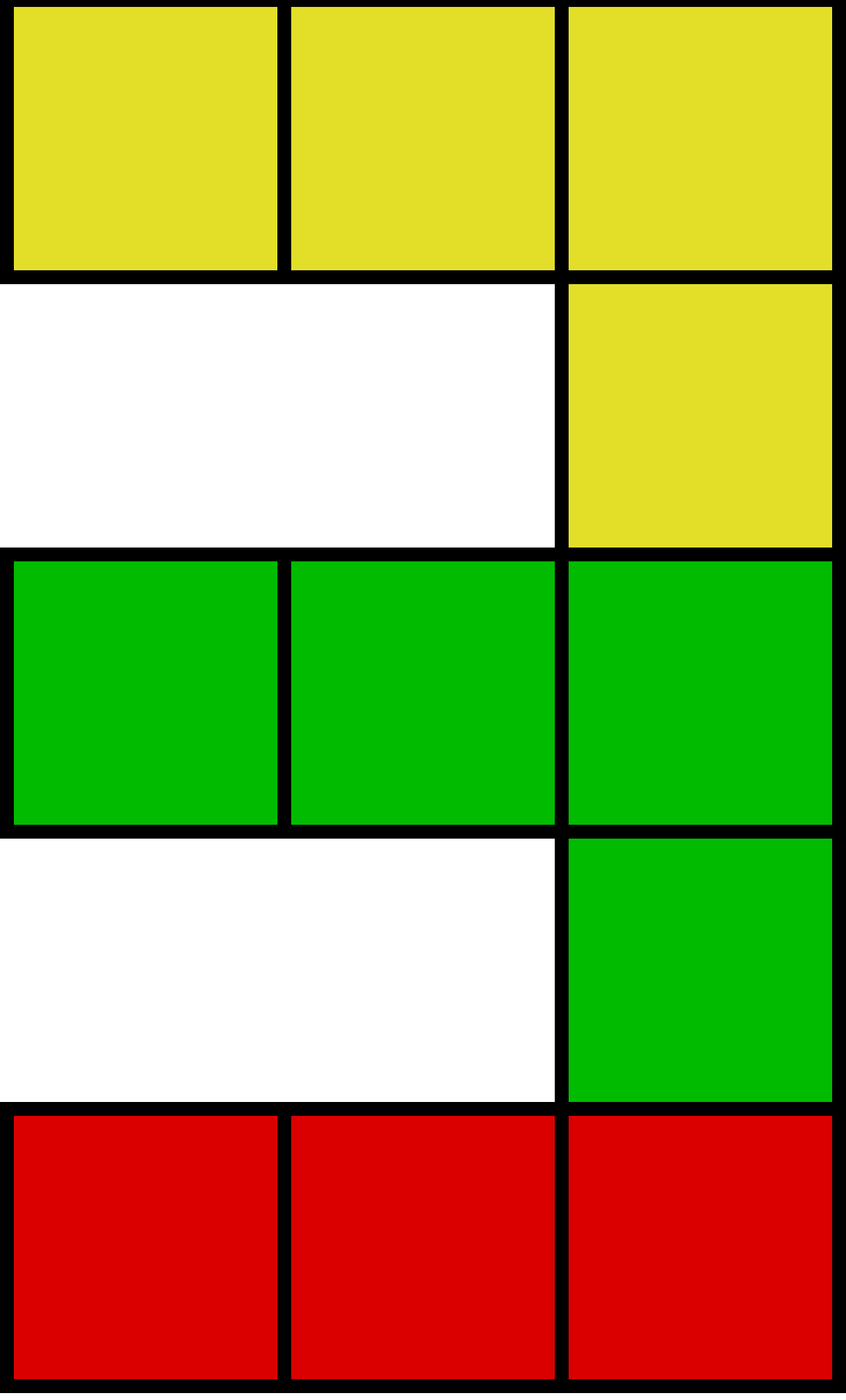}}
\newcommand{\rqfour}{\rqtag{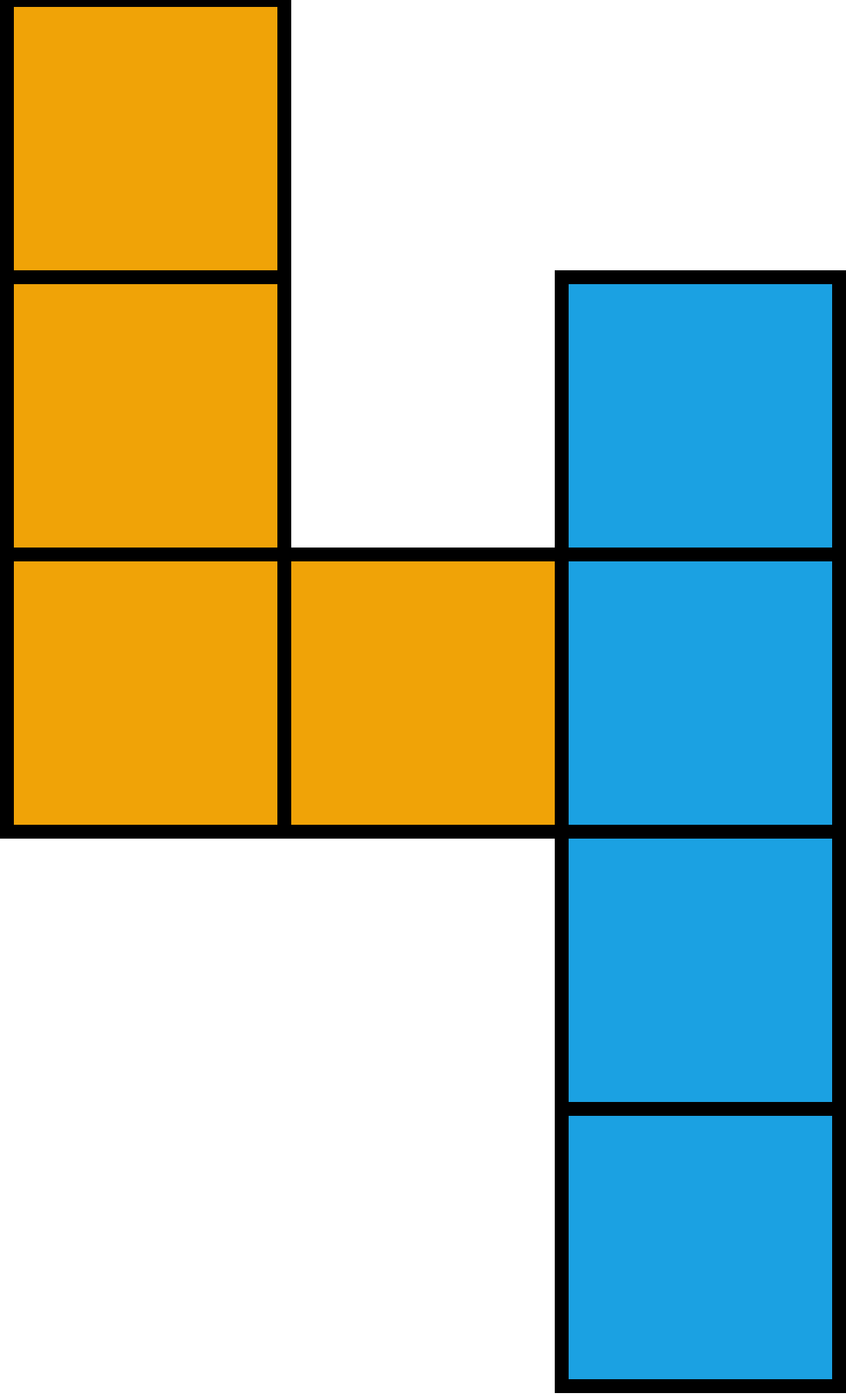}}
\renewcommand{\rq}[1]{\ifcase#1\relax\or\rqone\or\rqtwo\or\rqthree\or\rqfour\else\texttt{RQ-#1}\fi}

\newcommand{\tracepolicy}[1]{\textcolor{tracePolicy}{#1}}
\newcommand{\tracevalue}[1]{\textcolor{traceValue}{#1}}
\newcommand{\tracestate}[1]{\textcolor{traceState}{#1}}

\newcommand*{\cnum}[1]{%
  \tikz[baseline=(N.base)]{%
    \node[circle, fill=black, text=white, inner sep=0pt,
          minimum size=2ex, font=\scriptsize\sffamily\bfseries] (N) {#1};}}

\usepackage{threeparttable}
\newcommand{\thinkon}{\faLightbulbO}
\newcommand{\thinkoff}{%
  \makebox[1em][c]{\faLightbulbO}%
  \hspace{-1em}%
  \makebox[1em][c]{\raisebox{0.7ex}{\rotatebox[origin=c]{-45}{\rule{1.15em}{0.08em}}}}%
}

\IfFileExists{acro.sty}{%
  \DeclareAcronym{spsd}{
    short = SPSD,
    long = self-play search distillation,
  }%
  \newcommand{\methodtitle}{Self-Play Search Distillation\xspace}%
  \newcommand{\SPSD}{SPSD\xspace}%
}{%
  \newcommand{\methodtitle}{Self-Play Search Distillation\xspace}%
  \newcommand{\SPSD}{SPSD\xspace}%
}
\newcommand{\OPSD}{OPSD\xspace}
\newcommand{\projectedresult}[1]{\textcolor{black!62}{\ensuremath{\mathit{#1}}}}

\newcommand{\afficon}[1]{\includegraphics[width=7pt,height=7pt]{#1.png}}
\newcommand{\affmark}[1]{\hspace{1.5pt}\raisebox{0.5ex}{\afficon{#1}}}
\newcommand{\affinline}[1]{\raisebox{-0.1ex}{\afficon{#1}}\,}

\title{\methodtitle \\ for Large Language Model Reasoning}
\date{}
\author{%
Lorenzo Molfetta\affmark{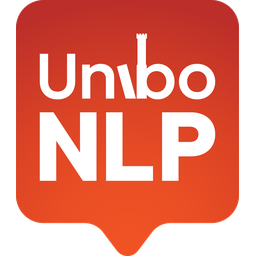} \quad Wai-Chung Kwan\affmark{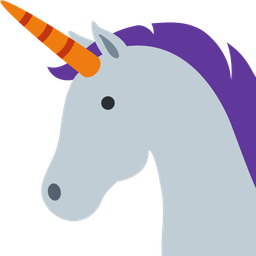} \quad Giacomo Frisoni\affmark{unibonlp} \quad Luca Ragazzi\affmark{unibonlp} \\
\textbf{Gianluca Moro}\affmark{unibonlp} \quad \textbf{Pavlos Vougiouklis}\affmark{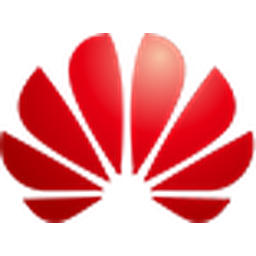} \quad \textbf{Jeff Z. Pan}\affmark{edinburgh}\affmark{huawei} \quad \textbf{Pasquale Minervini}\affmark{edinburgh}\affmark{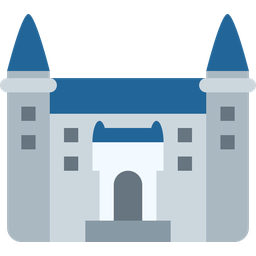} \\
\affinline{unibonlp}University of Bologna \qquad
\affinline{edinburgh}University of Edinburgh \\
\affinline{huawei}Huawei Technologies R\&D (UK) Ltd. \qquad
\affinline{miniml}Miniml.AI
\\
\texttt{lorenzo.molfetta@unibo.it}
}
\hypersetup{
  pdftitle={Self-Play Search Distillation for Large Language Model Reasoning},
  pdfauthor={Lorenzo Molfetta; Wai-Chung Kwan; Giacomo Frisoni; Luca Ragazzi; Gianluca Moro; Pavlos Vougiouklis; Jeff Z. Pan; Pasquale Minervini},
  pdfsubject={},
  pdfkeywords={}
}

\begin{document}
\maketitle

\begin{abstract}
Improving reasoning abilities in Large Language Models (LLMs) requires high-quality data that exposes difficult decisions, competing alternatives, and their consequences.
Data scarcity is driven by the low quality of synthetic data and the cost of human labeling.
We introduce Self-Play Search Distillation (\SPSD), a framework for generating superhuman synthetic data via self-play of MuZero-like networks trained on board games.
\SPSD\ uses executable environments to turn search into structured reasoning problems.
At each state, the expert identifies a preferred decision, plausible alternatives, plausible opponent replies, and value estimates.
By converting the self-play search records into superhuman chains-of-thought, we train LLMs with environment-grounded supervision.
Although trained only on self-play search records, \SPSD\ transfers to unseen mathematics.
On Qwen3-4B-Base, it raises the mean over six mathematics benchmarks from 24.1 to 36.6 while increasing the held-out-game win rate from 15\% to 45\%.
\SPSD\ offers an annotation-efficient way to create high-quality synthetic data for improving LLM performance in reasoning tasks.
\end{abstract}

\section{Introduction}
\label{sec:introduction}

Scaling the reasoning capabilities of large language models (LLMs) requires more than simply increasing the amount of training data.
It requires synthetic supervision whose quality can exceed the student's current capabilities.
Final answers reveal whether a solution succeeded, but not how the solver preserved state, rejected plausible alternatives, or anticipated delayed consequences~\citep{DBLP:conf/iclr/0002HBVPW23,DBLP:conf/blackboxnlp/NandaLW23,DBLP:conf/nips/ValmeekamMHSK23,DBLP:conf/nips/ValmeekamMSK23}.
Human process supervision can expose these operations, but it is costly to collect, difficult to verify, and ultimately bounded by the expertise of its annotators~\citep{DBLP:conf/nips/Ouyang0JAWMZASR22}.

Recent self-generation methods offer a scalable alternative by allowing the learner to construct tasks, traces, or feedback~\citep{DBLP:conf/icml/ChenDYJG24,DBLP:journals/corr/abs-2506-24119,DBLP:journals/corr/abs-2505-03335}.
Yet, when the learner itself participates in generating the supervision, data quality remains coupled to its current errors and limitations.
The central challenge is to generate structured supervision that exposes comparisons and consequences beyond the student's current policy and can be checked independently.

Executable board games provide a compact testbed for this challenge.
Their rules define exact transitions and legal actions, and terminal outcomes expose delayed consequences, while controlled rule changes vary the reasoning problem without changing the evaluation interface.
Search-based self-play systems in this lineage have achieved superhuman play and, crucially, retain more than the selected action: they expose alternatives, opponent replies, branch values, and replayable consequences~\citep{DBLP:journals/corr/abs-1712-01815,DBLP:journals/nature/SchrittwieserAH20,DBLP:conf/nips/YeLKAG21}.
A single root search can therefore provide a preferred action, a distribution over legal alternatives, branch values, and continuations that can be checked independently by replay.
These records capture the comparisons and delayed consequences absent from outcome-only game logs.
Crucially, search-based self-play has already demonstrated that such experts can surpass human performance, providing a source of supervision that is not inherently capped by human expertise~\citep{DBLP:journals/corr/abs-1712-01815,DBLP:journals/nature/SchrittwieserAH20}.
By placing these external search experts---rather than the LLM itself---in the data-generation loop, we decouple supervision quality from the student's current policy.
Self-play search thus becomes more than a game-playing mechanism.
It provides a programmable source of search-derived reasoning traces whose alternatives and consequences can be checked by replay under the executable rules of each environment.

We introduce Self-Play Search Distillation (SPSD), an offline bridge from superhuman-capable self-play search to LLM post-training.
Given a search expert, \SPSD converts replayable decision states into verifier-grounded supervision that exposes the selected action, plausible alternatives, and their consequences, together with complementary questions about the underlying state.
The LLM never participates in expert self-play during post-training and receives no search information at inference time.
Our experiments show that \SPSD can improve mathematical reasoning while preserving game competence, although the magnitude of the gains depends on the backbone.

We make three contributions.
\begin{enumerate}[leftmargin=*,noitemsep,topsep=2pt,labelsep=0.6em,label={\protect\cnum{\arabic*}}]
\item We introduce an offline framework that turns self-play search into a scalable source of replay-grounded supervision for LLM reasoning.
\item We develop a replay-grounded interface that converts expert decisions, alternatives, and state transitions into reasoning traces and six complementary state-supervision tasks.
\item We evaluate whether this supervision transfers beyond the source search environments, across game play, controlled rule variation, and mathematical reasoning.
\end{enumerate}

\begingroup
\definecolor{qwenPurple}{HTML}{6C4CE0}   %
\definecolor{gemmaBlue}{HTML}{1A73E8}    %
\definecolor{ossRed}{HTML}{D62728}       %
\definecolor{museTeal}{HTML}{0E8F8F}     %
\definecolor{opusOrange}{HTML}{D97757}   %
\tikzset{
  qwenThreeLine/.style={draw=qwenPurple},
  qwenThreeFiveLine/.style={draw=qwenPurple},
  gemmaLine/.style={draw=gemmaBlue},
  gptLine/.style={draw=ossRed},
  llamaLine/.style={draw=museTeal},
  opusLine/.style={draw=opusOrange},
}
\newcommand{\bandstretch}{3.5}
\pgfmathdeclarefunction{bandlog}{1}{%
  \begingroup
  \pgfkeys{/pgf/fpu=true,/pgf/fpu/output format=fixed}%
  \pgfmathparse{ln(#1)/ln(2)}%
  \pgfmathsetmacro{\bandl}{\pgfmathresult}%
  \pgfmathparse{\bandl<4 ? \bandl : (\bandl<6 ? 4+\bandstretch*(\bandl-4) : 4+2*\bandstretch+(\bandl-6))}%
  \pgfmathsmuggle\pgfmathresult
  \endgroup
}
\pgfmathdeclarefunction{bandloginv}{1}{%
  \begingroup
  \pgfkeys{/pgf/fpu=true,/pgf/fpu/output format=fixed}%
  \pgfmathparse{#1<4 ? 2^(#1) : (#1<4+2*\bandstretch ? 2^(4+(#1-4)/\bandstretch) : 2^(6+(#1-4-2*\bandstretch)))}%
  \pgfmathsmuggle\pgfmathresult
  \endgroup
}
\pgfplotsset{
  bandaxis/.style={
    x coord trafo/.code={\pgfmathparse{bandlog(##1)}},
    x coord inv trafo/.code={\pgfmathparse{bandloginv(##1)}},
  },
}
\newcommand{\sembaropacity}{0.55}
\newcommand{\sembarcap}{1.7pt}
\newcommand{\sembarwidth}{0.5pt}
\newcommand{\sembar}[5]{%
  \pgfplotsextra{%
    \pgfmathsetmacro{\sembarlo}{#2-#3}%
    \pgfmathsetmacro{\sembarhi}{#2+#3}%
    \begin{scope}[#4,line width=\sembarwidth,opacity=\sembaropacity,xshift=#5]
      \draw (axis cs:#1,\sembarlo) -- (axis cs:#1,\sembarhi);
      \draw ([xshift=-\sembarcap]axis cs:#1,\sembarlo) -- ([xshift=\sembarcap]axis cs:#1,\sembarlo);
      \draw ([xshift=-\sembarcap]axis cs:#1,\sembarhi) -- ([xshift=\sembarcap]axis cs:#1,\sembarhi);
    \end{scope}%
  }%
}
\newcommand{\plotpointfull}[7]{%
  \pgfplotsextra{%
    \begin{scope}[xshift=#5]
      \path (axis cs:#1,#2) node[inner sep=0pt] {\scalebox{#6}{#3}};
      \path (axis cs:#1,#2) node[anchor=#7,font=\fontsize{4.9}{4.9}\selectfont,
            text=black!70,scale=0.88,transform shape] {#4};
    \end{scope}%
  }%
}
\newcommand{\gamepoint}[7]{%
  \sembar{#1}{#2}{#3}{#4}{#7}%
  \plotpointfull{#1}{#2}{#5}{#6}{#7}{1}{north}%
}
\newcommand{\gamepointc}[9]{%
  \sembar{#1}{#2}{#3}{#4}{#7}%
  \plotpointfull{#1}{#2}{#5}{#6}{#7}{#8}{#9}%
}
\newcommand{\mathpoint}[6]{\plotpointfull{#1}{#2}{#3}{#4}{#5}{1}{north}}
\newcommand{\mathpointc}[7]{\plotpointfull{#1}{#2}{#3}{#4}{#5}{#6}{#7}}
\newcommand{\legendplain}[2]{\mbox{#1\hspace{0.7mm}#2}}
\newcommand{\qwenversion}[4]{%
  \pgfplotsextra{%
    \begin{scope}[xshift=#4]
    \path (axis cs:#1,#2) node[
      anchor=south,
      inner sep=0pt,
      font=\fontsize{4.15}{4.15}\selectfont\bfseries,
      text=qwenVersion,
      scale=0.78,
      transform shape,
      yshift=3.0pt
    ] {#3};%
    \end{scope}%
  }%
}
\newcommand{\gptvariant}[4]{%
  \pgfplotsextra{%
    \begin{scope}[xshift=#4]
    \path (axis cs:#1,#2) node[
      anchor=south,
      inner sep=0pt,
      font=\fontsize{5.0}{5.0}\selectfont\bfseries,
      text=ossRed,
      scale=0.78,
      transform shape,
      yshift=5.5pt
    ] {#3};%
    \end{scope}%
  }%
}
\newcommand{\gptvariantxy}[5]{%
  \pgfplotsextra{%
    \begin{scope}[xshift=#4]
    \path (axis cs:#1,#2) node[
      anchor=south,
      inner sep=0pt,
      font=\fontsize{5.0}{5.0}\selectfont\bfseries,
      text=ossRed,
      scale=0.78,
      transform shape,
      yshift=#5
    ] {#3};%
    \end{scope}%
  }%
}
\newcommand{\apivariant}[6]{%
  \pgfplotsextra{%
    \begin{scope}[xshift=#4]
    \path (axis cs:#1,#2) node[
      anchor=south,
      inner sep=0pt,
      font=\fontsize{5.0}{5.0}\selectfont\bfseries,
      text=#6,
      scale=0.78,
      transform shape,
      yshift=#5
    ] {#3};%
    \end{scope}%
  }%
}
\pgfplotsset{
  rqaxis/.style={
    width=0.53\textwidth,
    height=5.25cm,
    bandaxis,
    xmin=0.33,
    xmax=34000,
    xtick={0.6,1,2,4,8,16,32,64,128,400,4800},
    xticklabels={0.6\hspace{3pt},\hspace{3pt}1,2,4,8,16,32,64\hspace{2.5pt},\hspace{4pt}128,400,?},
    grid=both,
    major grid style={dashed,gray!45},
    minor grid style={dashed,gray!28},
    minor y tick num=0,
    axis background/.style={fill=rqplotbg},
    every tick label/.append style={font=\fontsize{6.8}{6.8}\selectfont},
    xlabel={\texttt{model size (B)}},
    xlabel style={font=\fontsize{7.4}{7.4}\selectfont,yshift=2pt},
    title style={font=\fontsize{7.8}{7.8}\selectfont,yshift=-1ex},
    clip=false,
  },
}
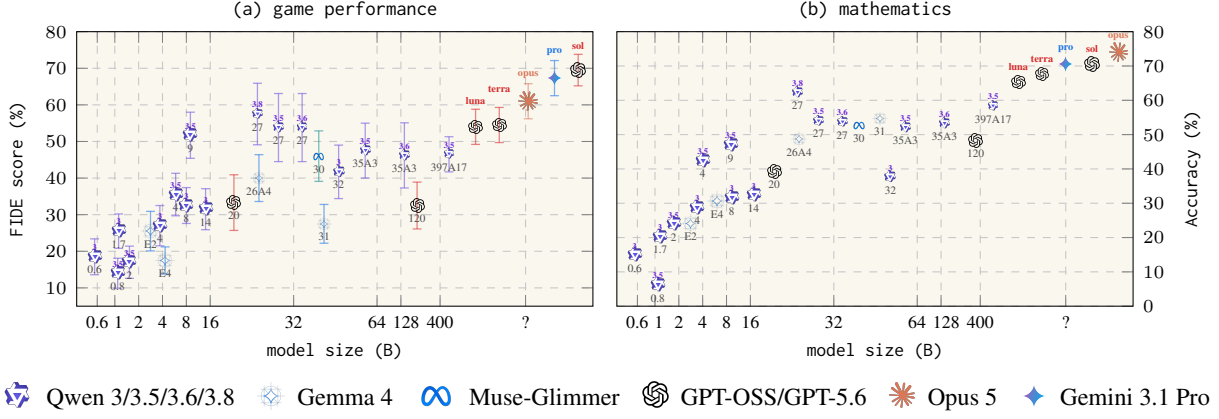
\begin{figure*}[t]
    \centering
    \resizebox{\textwidth}{!}{%
    \begin{tikzpicture}
        \begin{axis}[
            rqaxis,
            name=games,
            ymin=5,
            ymax=80,
            ytick={10,20,30,40,50,60,70,80},
            ylabel={\texttt{FIDE score (\%)}},
            ylabel style={font=\fontsize{7.4}{7.4}\selectfont},
            title={\texttt{(a) game performance}},
        ]
            \gamepoint{0.6}{18.5}{4.9}{qwenThreeLine}{\qwenploticon}{0.6}{-1pt}
            \qwenversion{0.6}{18.5}{3}{-1pt}
            \gamepoint{0.8}{14.0}{4.1}{qwenThreeFiveLine}{\qwenploticon}{0.8}{4pt}
            \qwenversion{0.8}{14.0}{3.5}{4pt}
            \gamepoint{1.7}{25.5}{4.7}{qwenThreeLine}{\qwenploticon}{1.7}{-5.5pt}
            \qwenversion{1.7}{25.5}{3}{-5.5pt}
            \gamepoint{2}{25.5}{5.4}{gemmaLine}{\gemmaploticon}{E2}{4.5pt}
            \gamepoint{2}{17.0}{4.4}{qwenThreeFiveLine}{\qwenploticon}{2}{-3.5pt}
            \qwenversion{2}{17.0}{3.5}{-3.5pt}
            \gamepoint{4}{27.0}{5.5}{qwenThreeLine}{\qwenploticon}{4}{-1pt}
            \qwenversion{4}{27.0}{3}{-1pt}
            \gamepoint{4}{35.5}{5.8}{qwenThreeFiveLine}{\qwenploticon}{4}{5pt}
            \qwenversion{4}{35.5}{3.5}{5pt}
            \gamepoint{4}{17.5}{3.7}{gemmaLine}{\gemmaploticon}{E4}{1pt}
            \gamepoint{8}{32.5}{4.9}{qwenThreeLine}{\qwenploticon}{8}{0pt}
            \qwenversion{8}{32.5}{3}{0pt}
            \gamepoint{9}{51.7}{6.3}{qwenThreeFiveLine}{\qwenploticon}{9}{0pt}
            \qwenversion{9}{51.7}{3.5}{0pt}
            \gamepoint{14}{31.5}{5.6}{qwenThreeLine}{\qwenploticon}{14}{0pt}
            \qwenversion{14}{31.5}{3}{0pt}
            \gamepointc{20}{33.3}{7.6}{gptLine}{\openaiploticon}{20}{-1.2pt}{0.78}{north}
            \gamepointc{26}{40.0}{6.4}{gemmaLine}{\gemmaploticon}{26A4}{-3.7pt}{0.78}{north}
            \gamepointc{27}{57.5}{8.4}{qwenThreeLine}{\qwenploticon}{27}{-6pt}{0.72}{north}
            \qwenversion{27}{57.5}{3.8}{-6pt}
            \gamepointc{27}{53.8}{9.3}{qwenThreeFiveLine}{\qwenploticon}{27}{2pt}{0.72}{north}
            \qwenversion{27}{53.8}{3.5}{2pt}
            \gamepointc{27}{53.8}{9.3}{qwenThreeLine}{\qwenploticon}{27}{11pt}{0.72}{north}
            \qwenversion{27}{53.8}{3.6}{11pt}
            \gamepointc{30}{46.0}{6.9}{llamaLine}{\llamaploticon}{30}{12.5pt}{0.78}{north}
            \gamepointc{31}{27.5}{5.3}{gemmaLine}{\gemmaploticon}{31}{13pt}{0.78}{north}
            \gamepointc{32}{41.7}{7.3}{qwenThreeLine}{\qwenploticon}{32}{17.1pt}{0.78}{north}
            \qwenversion{32}{41.7}{3}{17.1pt}
            \gamepointc{35}{47.5}{7.5}{qwenThreeFiveLine}{\qwenploticon}{35A3}{23.1pt}{0.72}{north}
            \qwenversion{35}{47.5}{3.5}{23.1pt}
            \gamepointc{35}{46.2}{8.9}{qwenThreeLine}{\qwenploticon}{35A3}{38pt}{0.72}{north}
            \qwenversion{35}{46.2}{3.6}{38pt}
            \gamepointc{120}{32.5}{6.4}{gptLine}{\openaiploticon}{120}{7pt}{0.78}{north}
            \gamepointc{397}{46.5}{4.8}{qwenThreeFiveLine}{\qwenploticon}{397A17}{3.5pt}{0.70}{north}
            \qwenversion{397}{46.5}{3.5}{3.5pt}
            \gamepointc{4800}{54.0}{4.8}{gptLine}{\openaiploticon}{}{-19pt}{0.75}{north}
            \gptvariantxy{4800}{58.8}{luna}{-19pt}{2.5pt}
            \gamepointc{4800}{54.5}{4.8}{gptLine}{\openaiploticon}{}{-10pt}{0.75}{north}
            \gptvariantxy{4800}{59.3}{terra}{-10pt}{7.5pt}
            \gamepointc{4800}{61.0}{4.8}{opusLine}{\claudeploticon}{}{1pt}{1}{north}
            \apivariant{4800}{65.8}{opus}{1pt}{2.5pt}{opusOrange}
            \gamepointc{4800}{67.31}{4.8}{gemmaLine}{\geminiploticon}{}{11pt}{1}{north}
            \apivariant{4800}{72.11}{pro}{11pt}{2.5pt}{gemmaBlue}
            \gamepointc{4800}{69.5}{4.3}{gptLine}{\openaiploticon}{}{20pt}{0.85}{north}
            \gptvariantxy{4800}{73.8}{sol}{20pt}{2.5pt}
        \end{axis}

        \begin{axis}[
            rqaxis,
            name=math,
            at={(games.east)},
            xshift=0.32cm,
            anchor=west,
            ymin=0,
            ymax=80,
            ytick={0,10,20,30,40,50,60,70,80},
            ytick pos=right,
            ylabel={\texttt{Accuracy (\%)}},
            ylabel style={font=\fontsize{7.4}{7.4}\selectfont,at={(axis description cs:1.15,0.5)},anchor=south},
            title={\texttt{(b) mathematics}},
        ]
            \mathpoint{0.6}{14.80}{\qwenploticon}{0.6}{-1pt}{0pt}
            \qwenversion{0.6}{14.80}{3}{-1pt}
            \mathpoint{0.8}{6.15}{\qwenploticon}{0.8}{4pt}{0pt}
            \qwenversion{0.8}{6.15}{3.5}{4pt}
            \mathpoint{1.7}{20.20}{\qwenploticon}{1.7}{-5pt}{0pt}
            \qwenversion{1.7}{20.20}{3}{-5pt}
            \mathpoint{2}{23.85}{\qwenploticon}{2}{-2pt}{0pt}
            \qwenversion{2}{23.85}{3.5}{-2pt}
            \mathpoint{2}{24.05}{\gemmaploticon}{E2}{4.5pt}{0pt}
            \mathpoint{4}{28.65}{\qwenploticon}{4}{-2pt}{0pt}
            \qwenversion{4}{28.65}{3}{-2pt}
            \mathpoint{4}{30.65}{\gemmaploticon}{E4}{5.5pt}{0pt}
            \mathpoint{4}{42.45}{\qwenploticon}{4}{0pt}{0pt}
            \qwenversion{4}{42.45}{3.5}{0pt}
            \mathpoint{8}{31.55}{\qwenploticon}{8}{2pt}{0pt}
            \qwenversion{8}{31.55}{3}{2pt}
            \mathpoint{9}{46.95}{\qwenploticon}{9}{0pt}{0pt}
            \qwenversion{9}{46.95}{3.5}{0pt}
            \mathpoint{14}{32.35}{\qwenploticon}{14}{3pt}{0pt}
            \qwenversion{14}{32.35}{3}{3pt}
            \mathpointc{20}{39.30}{\openaiploticon}{20}{-1.2pt}{0.78}{north}
            \mathpointc{26}{48.75}{\gemmaploticon}{26A4}{-3.7pt}{0.78}{north}
            \mathpointc{27}{62.20}{\qwenploticon}{27}{-6pt}{0.72}{north}
            \qwenversion{27}{62.20}{3.8}{-6pt}
            \mathpointc{27}{54.00}{\qwenploticon}{27}{2pt}{0.72}{north}
            \qwenversion{27}{54.00}{3.5}{2pt}
            \mathpointc{27}{53.65}{\qwenploticon}{27}{11pt}{0.72}{north}
            \qwenversion{27}{53.65}{3.6}{11pt}
            \mathpointc{30}{52.75}{\llamaploticon}{30}{12.5pt}{0.78}{north}
            \mathpointc{31}{54.65}{\gemmaploticon}{31}{19pt}{0.78}{north}
            \mathpointc{32}{37.55}{\qwenploticon}{32}{21.5pt}{0.78}{north}
            \qwenversion{32}{37.55}{3}{21.5pt}
            \mathpointc{35}{52.10}{\qwenploticon}{35A3}{23.1pt}{0.72}{north}
            \qwenversion{35}{52.10}{3.5}{23.1pt}
            \mathpointc{35}{53.15}{\qwenploticon}{35A3}{38pt}{0.72}{north}
            \qwenversion{35}{53.15}{3.6}{38pt}
            \mathpointc{120}{48.35}{\openaiploticon}{120}{14pt}{0.78}{north}
            \mathpointc{397}{58.30}{\qwenploticon}{397A17}{5pt}{0.70}{north}
            \qwenversion{397}{58.30}{3.5}{5pt}
            \mathpointc{4800}{65.30}{\openaiploticon}{}{-18pt}{0.75}{north}
            \gptvariant{4800}{65.30}{luna}{-18pt}
            \mathpointc{4800}{67.70}{\openaiploticon}{}{-9pt}{0.75}{north}
            \gptvariant{4800}{67.70}{terra}{-9pt}
            \mathpoint{4800}{70.55}{\geminiploticon}{}{0pt}{0pt}
            \apivariant{4800}{70.55}{pro}{0pt}{5.5pt}{gemmaBlue}
            \mathpointc{4800}{70.65}{\openaiploticon}{}{10pt}{0.85}{north}
            \gptvariant{4800}{70.65}{sol}{10pt}
            \mathpoint{4800}{74.05}{\claudeploticon}{}{20pt}{0pt}
            \apivariant{4800}{74.05}{opus}{20pt}{5.5pt}{opusOrange}
        \end{axis}
    \end{tikzpicture}%
    }
    \resizebox{\textwidth}{!}{\mbox{\fontsize{7.4}{7.4}\selectfont
    \legendplain{\qwenicon}{Qwen 3/3.5/3.6/3.8} \hspace{1.1mm}
    \legendplain{\gemmaicon}{Gemma 4} \hspace{1.1mm}
    \legendplain{\llamaicon}{Muse-Glimmer} \hspace{1.1mm}
    \legendplain{\openaiicon}{GPT-OSS/GPT-5.6} \hspace{1.1mm}
    \legendplain{\claudeicon}{Opus 5} \hspace{1.1mm}
    \legendplain{\geminiicon}{Gemini 3.1 Pro}}}

    \caption{\textbf{Game and mathematics performance across model scale.} Panel (a) reports episode-mean FIDE with SEM whiskers on the SPSD board-game benchmark; panel (b) averages GPQA-Diamond and Humanity's Last Exam. While game competence tracks mathematical accuracy across scale, even frontier models do not saturate on board-game reasoning.}
    \label{fig:main-results}
\end{figure*}
\endgroup

\section{Related Work} \label{sec:related-work}

\paragraph{Game reasoning.}
Executable games support both the evaluation of stateful reasoning and the generation of interaction-grounded supervision.
ChessGPT~\cite{DBLP:conf/nips/FengLWTYSM0W23} connects policy learning with language modeling, while ChessBench~\cite{DBLP:journals/corr/abs-2402-04494} evaluates chess prediction against engine-derived targets.
Training on complete games~\cite{DBLP:conf/naacl/ZhangHLCL25} examines whether full-game context improves chess play, latent-state analyses~\cite{DBLP:journals/corr/abs-2403-15498} probe internal board representations, and chess reasoning testbeds~\cite{DBLP:conf/naacl/WangJWZLHW25} evaluate state tracking and tactical reasoning.
GTBench~\cite{DBLP:journals/corr/abs-2402-12348} evaluates strategic reasoning across game-theoretic settings, while lmgame-Bench~\cite{DBLP:journals/corr/abs-2505-15146} studies interactive play and transfer from game training.
KORGym~\cite{DBLP:journals/corr/abs-2505-14552} and TextArena~\cite{DBLP:journals/corr/abs-2504-11442} provide multi-turn game collections that support evaluation and training.
LLM-guided MCTS~\cite{DBLP:journals/corr/abs-2403-05632} uses language models to prune actions and estimate state values, while \citet{DBLP:conf/icml/SchultzAJLKPHSL25} compares explicit MCTS with search generated inside the language model.
While these approaches primarily evaluate game competence, train game-playing policies, or place the language model inside the planning loop, we train a search expert separately and let it export environment-grounded root decisions, alternatives, and continuations as offline language-model supervision.

\paragraph{Synthetic supervision.}
Synthetic methods derive supervision from verifiable tasks and explicit solution processes.
AlphaGeometry~\cite{DBLP:journals/nature/TrinhWLHL24} generates synthetic theorems and proofs without human demonstrations, Enigmata~\cite{DBLP:journals/corr/abs-2505-19914} pairs controllable puzzles with rule-based verifiers, and Absolute Zero~\cite{DBLP:journals/corr/abs-2505-03335} lets one model propose and solve executable reasoning tasks under programmatic feedback.
LogicPuzzleRL~\cite{DBLP:journals/corr/abs-2506-04821} trains with verifier rewards on logic puzzles, while Reasoning Gym~\cite{DBLP:journals/corr/abs-2505-24760} provides procedurally generated reasoning environments with verifiable rewards.
Complementary lines of work expose intermediate computation rather than only final outcomes.
Chain-of-thought prompting~\cite{DBLP:conf/nips/Wei0SBIXCLZ22} elicits stepwise rationales, Tree of Thoughts~\cite{DBLP:conf/nips/YaoYZS00N23} searches over alternative thoughts at inference, and ASTRO~\cite{DBLP:journals/corr/abs-2507-00417} trains on natural-language MCTS traces with reflection and backtracking.
Procedure cloning~\cite{DBLP:conf/nips/YangSAN22} imitates expert computations, while Stream of Search~\cite{DBLP:journals/corr/abs-2404-03683} bootstraps from flattened search trajectories.
RLoT~\cite{DBLP:journals/corr/abs-2505-14140} learns adaptive reasoning control through inference-time reinforcement learning, and ToTRL~\cite{DBLP:journals/corr/abs-2505-12717} reinforces tree-structured reasoning through puzzle solving.
Algorithm Distillation~\cite{DBLP:conf/iclr/LaskinWOPSSSHFB23} models learning histories to induce in-context policy improvement.
Studies of Othello sequence models~\cite{DBLP:conf/iclr/0002HBVPW23,DBLP:conf/blackboxnlp/NandaLW23} probe latent board-state representations, while PlanBench~\cite{DBLP:conf/nips/ValmeekamMHSK23} and a related critical planning evaluation~\cite{DBLP:conf/nips/ValmeekamMSK23} test plan generation and reasoning about change.
\SPSD also transfers structured decision information, but it supervises one root decision from a planning expert.
It retains ranked alternatives and replayable continuations that justify the selected action without exposing the full search tree or asking the student to reproduce the planner's internal computations.

\paragraph{Self-play and post-training.}
Language-model post-training draws supervision from people, teachers, model generations, and executable environments.
RLHF~\cite{DBLP:conf/nips/ChristianoLBMLA17,DBLP:conf/nips/Ouyang0JAWMZASR22} established preference-based post-training, DPO~\cite{DBLP:conf/nips/RafailovSMMEF23} optimizes preference pairs directly, and knowledge distillation~\cite{DBLP:journals/corr/HintonVD15} transfers predictive behavior from a teacher.
Self-Instruct~\cite{DBLP:conf/acl/WangKMLSKH23} bootstraps instruction data from model generations, while SPIN~\cite{DBLP:conf/icml/ChenDYJG24} iteratively contrasts target responses with responses produced by earlier model versions.
Beyond Human Data~\cite{DBLP:journals/tmlr/SinghCAAPGLH0XP24} filters self-generated solutions using binary feedback, DeepSeekMath~\cite{DBLP:journals/corr/abs-2402-03300} introduces GRPO for mathematical reasoning, ProRL~\cite{DBLP:journals/corr/abs-2505-24864} studies prolonged reinforcement learning, and VersaPRM~\cite{DBLP:journals/corr/abs-2502-06737} learns process rewards from synthetic reasoning data.
SPIRAL~\cite{DBLP:journals/corr/abs-2506-24119} directly optimizes an LLM through online multi-turn game self-play.
OpenSIR~\cite{kwan2025opensir} alternates a single policy between proposing and solving mathematical problems under difficulty- and diversity-calibrated rewards, and SCOPE~\cite{kwan2026scope} extends self-play to open-ended tasks by co-evolving a task-proposing challenger with a retrieval-augmented solver, graded by rubrics from a judge model.
Code-based self-play~\cite{DBLP:conf/ijcai/BachrachTHMJJFR25} refines executable game strategies through competition, while SPADE~\cite{liu2026spadeselfplayadaptivesynthetic} co-evolves executable tasks and a reasoning agent through hint-based regret.
ALE~\cite{DBLP:journals/jair/BellemareNVB13}, TextWorld~\cite{DBLP:conf/ijcai/CoteKYKBFMHAATT18}, Procgen~\cite{DBLP:conf/icml/CobbeHHS20}, Ludii~\cite{DBLP:conf/ecai/PietteSSSWB20}, and OpenSpiel~\cite{DBLP:journals/corr/abs-1908-09453} provide automatically scored settings for interaction and training.
POET~\cite{DBLP:journals/corr/abs-1901-01753}, unsupervised environment design~\cite{DBLP:conf/nips/DennisJVBRCL20}, and regret-based environment design~\cite{DBLP:conf/icml/Parker-HolderJ022} instead adapt the task distribution as the learner changes.
In \SPSD, self-play is part of corpus construction.
The decisions of the trained search experts are converted into an offline corpus.
This shifts self-play from optimizing the learner to producing reusable, environment-grounded reasoning data.

\section{Background}
\label{sec:background}

\paragraph{Expert Iteration.}
Expert Iteration (ExIt) alternates a planning expert that improves decisions at individual states with an apprentice that generalises those improvements across states~\citep{DBLP:conf/nips/AnthonyTB17,Anthony2021ExpertIteration}.
At iteration \(k\), an apprentice \(F_{\phi_k}\) guides expert search over a set of states \(\mathcal{D}_k\).
For each searched state, the planner produces a search-improved policy \tracepolicy{\(\pi_k^E\)}, and the training loop supplies an optional value target \tracevalue{\(\hat v_k\)}.
The next apprentice fits these targets through
\begin{equation}
\begin{aligned}
\mathcal{L}_k^{\mathrm{ExIt}}(\phi)
&=\sum_{s\in\mathcal{D}_k}
\Bigl[
\ell_\pi\!\left(F_\phi^\pi(\cdot\mid s),\tracepolicy{\pi_k^E(\cdot\mid s)}\right)\\
&\qquad\qquad+
\lambda_v\ell_v\!\left(F_\phi^v(s),\tracevalue{\hat v_k(s)}\right)
\Bigr],\\
\phi_{k+1}
&=\arg\min_\phi\mathcal{L}_k^{\mathrm{ExIt}}(\phi).
\end{aligned}
\label{eq:exit-apprentice-update}
\end{equation}
Here, \(\ell_\pi\) and \(\ell_v\) fit the policy and value targets, while \(\lambda_v=0\) yields policy-only training.
The value target may be a planner estimate, an observed return, or a bootstrapped target, depending on the ExIt instance.
Search concentrates computation on the current state, and apprentice fitting propagates the resulting improvements to later states.
Repeating these operations forms an approximate modified-policy-iteration loop~\citep{DBLP:journals/corr/abs-1904-03646,Anthony2021ExpertIteration}.
After training, the apprentice-guided planner can be queried for search-improved policies and values.

\paragraph{Neural-guided MCTS.}
Monte Carlo tree search provides a common planning mechanism within ExIt.
AlphaGo~\citep{DBLP:journals/nature/SilverHMGSDSAPL16} established neural-guided MCTS at scale.
AlphaGo Zero~\citep{DBLP:journals/nature/SilverSSAHGHBLB17} and AlphaZero~\citep{DBLP:journals/corr/abs-1712-01815} coupled this search process to iterative self-play policy and value fitting.
MuZero~\citep{DBLP:journals/nature/SchrittwieserAH20} performs the search through learned latent dynamics, and EfficientZero~\citep{DBLP:conf/nips/YeLKAG21} adds sample-efficiency mechanisms around the same planning interface.
For a game \(g\), let \(\mathcal{A}_g(s)\) denote the legal actions at state \(s\).
At ExIt iteration \(k\), let \tracevalue{\(Q_k(s,a)\)} denote the current action-value estimate at node \tracestate{\(s\)}, \tracepolicy{\(P_k(s,a)\)} the apprentice prior assigned at expansion, and \(N_k(s,a)\) the number of simulations selecting action \(a\).
Writing \(N_{k,s}=\sum_{b\in\mathcal{A}_g(\tracestate{s})}N_k(\tracestate{s},b)\), MuZero-style PUCT combines the backed-up value with a prior-weighted exploration bonus
\begin{equation}
\begin{aligned}
a_{\mathrm{tree}}
&=\arg\max_a\bigl[\tracevalue{Q_k(\tracestate{s},a)}+U_k(\tracestate{s},a)\bigr],\\
U_k(\tracestate{s},a)
&=\tracepolicy{P_k(\tracestate{s},a)}
\frac{\sqrt{N_{k,s}}}{1+N_k(\tracestate{s},a)}\\
&\quad\times\left(c_1+\log\frac{N_{k,s}+c_2+1}{c_2}\right).
\end{aligned}
\label{eq:puct}
\end{equation}
The value term favours actions with strong backed-up estimates, while the prior and visit-dependent bonus allocate simulations toward promising underexplored actions.
After a fixed simulation budget, root visits define the search-improved policy
\begin{equation}
\tracepolicy{\pi_k^E}(a\mid \tracestate{s})=
\frac{N_k(\tracestate{s},a)^{1/T_{\mathrm{mcts}}}}
{\sum_{b\in\mathcal{A}_g(\tracestate{s})}N_k(\tracestate{s},b)^{1/T_{\mathrm{mcts}}}}.
\label{eq:root-policy}
\end{equation}
The temperature \(T_{\mathrm{mcts}}\) controls the sharpness of Equation~\ref{eq:root-policy}, which supplies the policy target in Equation~\ref{eq:exit-apprentice-update}.
At \(T_{\mathrm{mcts}}=1\), it recovers the normalized root-visit target used in the original ExIt formulation~\citep{DBLP:conf/nips/AnthonyTB17}.
LightZero~\citep{DBLP:conf/nips/NiuPYLZRHLL23} provides a unified implementation of this neural-guided search family across diverse environments.
A completed root search also yields a selected action and a root value estimate, while retained paths can preserve concrete continuations behind these summaries.
These outputs capture both the expert decision and the evidence accumulated while comparing legal actions.

\paragraph{Search-derived supervision.}
The outputs of a completed search retain different levels of decision information.
A selected action records the final choice, the root distribution preserves relative support over legal actions, and retained continuations expose the state changes and delayed consequences behind that comparison.
Reasoning chains and procedure-cloning targets can convert this intermediate structure into direct supervision~\citep{DBLP:conf/nips/Wei0SBIXCLZ22,DBLP:conf/nips/YangSAN22,DBLP:journals/corr/abs-2404-03683}.
Executable environments provide an additional guarantee, because each narrated transition or outcome can be checked against them.
On-policy self-distillation (OPSD) further allows privileged evidence to guide a teacher along prefixes sampled by the current student~\citep{DBLP:journals/corr/abs-2601-19897}.
Together, these ingredients motivate an intermediate representation that preserves search evidence in executable form before any choice of language form or post-training objective.

\section{Method}
\label{sec:method}

\SPSD turns this intermediate representation into an offline bridge from self-play search to LLM post-training.
\SPSD comprises three stages.
A search expert plays against itself and exports one record for each decision state it reaches. Each record is then replayed in the environment, and records whose fields cannot be reproduced are discarded. The surviving records are converted into prompts and reference responses.
The expert is held fixed during generation, so the supervision it produces is independent of the student trained on it.
For each source game we train an ExIt-family search expert~\citep{DBLP:conf/nips/AnthonyTB17} and select a training step, suppressing the ExIt iteration index to write the search outputs as \(\pi^E\), \(v^E\), and \(\mathcal{B}^E\).

\begin{figure*}[t]
\centering
\includegraphics[width=\textwidth]{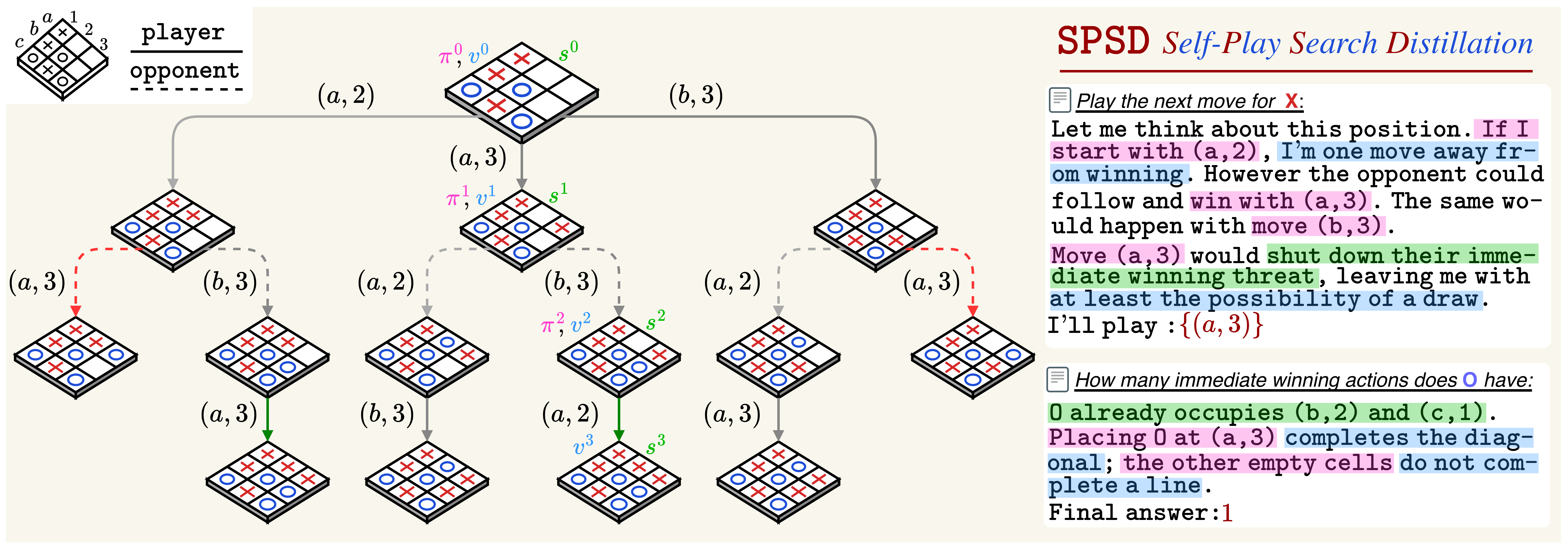}
\caption{\textbf{From self-play search to language supervision.} A search expert exports selected actions, root-level estimates, and replayable continuations.
These records are validated before they become move-choice and state-question supervision for language-model post-training.}
\label{fig:main-method}
\end{figure*}

\subsection{Executable decision records}
\label{subsec:record-interface}

\paragraph{Game interface.}
Each game \(g\) defines a state space \(\mathcal{S}_g\), legal-action function \(\mathcal{A}_g(s)\), transition function \(T_g(s,a)\), terminal and outcome functions, and state formatter \(\rho_g\).
The executable functions determine which states and transitions are valid, while \(\rho_g\) maps a restored state into the shared language interface.

\paragraph{Expert record.}
The expert record stores the selected decision together with the search evidence needed to construct its supervision.
To enforce diversity in the data while accounting for the deterministic nature of the expert, we begin each trajectory with a uniformly random legal prefix and thereby distribute retained decisions across the reachable state space.
The expert controls both seats after the prefix.
At each selected state, it runs 50 MCTS simulations, selects the root action \(a^\star(s)\), and exports
\begin{equation}
m(s)=
\bigl(
 s,
 \mathcal{A}_g(s),
 a^\star(s),
 \tracepolicy{\pi^E(\cdot\mid s)},
 \tracevalue{v^E(s)},
 \mathcal{B}^E(s)
\bigr).
\label{eq:expert-record}
\end{equation}
where \(\mathcal{A}_g(s)\) is the legal-action set at \(s\), \(a^\star(s)\) the selected root action, \(\pi^E(\cdot\mid s)\) the visit-derived root policy, \(v^E(s)\) the root value estimate, and \(\mathcal{B}^E(s)\) the retained branch evidence.
For each exported root action, \(\mathcal{B}^E(s)\) stores its visit count, export index, edge value or missing-value marker, and replayable prefix.
Each retained branch is replayed in the environment from the prefix stored in \(\mathcal{B}^E(s)\), recovering its successor states and any terminal outcome, and verbalized from those transitions.

\subsection{Replay-grounded supervision}
\label{subsec:grounded-supervision}

\paragraph{Supervision interface.}
Each decision record becomes a replay-grounded supervision training instance.
Materialization produces the row \((x,y;e)\), where \(x\) is the board-game state, \(y\) is the target best action, and \(e\) is the target-hidden expert trace derived from its MCTS process.
The prompt is built from \(\rho_g(s)\) and includes the rules, board, side to move, the list of legal options, and the answer contract.
Both \(y\) and \(e\) come from the verified contents of \(m(s)\).
The student receives only \(x\), so it must reconstruct the comparison from the board and the legal-action list.

\paragraph{Move choice.}
For move-choice rows, \tracepolicy{\(\pi^E\)} and \(\mathcal{B}^E\) rank the legal root actions.
Replay follows the expert-selected action, up to one visit-ranked legal alternative, and, when available, one opponent reply from the selected continuation.
Each retained branch contains its ordered action prefix, recomputed successor states, and a terminal outcome upon reaching a terminal state.
The generated trace verbalizes only the board effects and outcomes recovered from these transitions.
Linearization preserves three invariants.
The final handle remains \(a^\star(s)\), every narrated prefix replays legally, and equivalent failed branches are deduplicated.
Terminal outcomes appear only when replay reaches a terminal state, while non-terminal comparisons remain anchored to the retained root continuations.
The trace concludes with the expert-selected boxed legal handle.
\Cref{app:grounded-linearization} details the linearization algorithm and record schema, and \cref{app:grounded-mcts-example} shows a linearized record.

\paragraph{State questions.}
A model can select a strong move while misreading the board that justifies it.
To supervise the reading itself, \SPSD adds six environment-derived question types covering occupancy, legality, threat count, legal-action count, legal-action enumeration, and successor state.
At every retained decision state, the answer to the selected state question is computed from the board, legal-action set, transition function, and terminal rules.
Across games, \SPSD applies the same executable interface to all six question types. The prompt format remains fixed, while each answer is computed from the rules and transitions of the selected game.
Splits are grouped by source trajectory, and each decision state belongs to exactly one row family, which prevents a state from being both trained and probed.
\Cref{fig:main-method} illustrates an example of move-choice reasoning and a state question derived from an expert MCTS record.
\Cref{app:spsd-dataset-taxonomy,app:materialization} report the question types, row-construction procedure, and privileged fields.

\subsection{Language-model post-training}
\label{subsec:optimization}

We consider three post-training paths for distilling expert search into the LLMs.
\paragraph{Supervised Fine-Tuning.}
SFT trains the language model directly on \((e,y)\), exposing the replay-grounded response as the target sequence.

\paragraph{On-policy self-distillation.}
SFT is fitted to states the expert visits, which are not the states a weaker student reaches.
To align teacher guidance with the states induced by the student's own responses while retaining the expert trace as privileged evidence, we employ \OPSD~\citep{DBLP:journals/corr/abs-2601-19897}, which lets the student sample its own response \(y=(y_1,\ldots,y_T)\) from \(p_\theta(\cdot\mid x)\) while the trace \(e\) conditions the teacher rather than the student.
A fixed teacher with parameters \(\bar\theta\) evaluates the same student-generated prefix while conditioning on the expert trace
\begin{equation}
q_{\bar\theta}^{e}(\cdot\mid x,y_{<t})
=
p_{\bar\theta}(\cdot\mid x,e,y_{<t}).
\label{eq:opsd-teacher}
\end{equation}
where \(\bar\theta\) are the frozen teacher parameters, \(e\) the privileged trace of the current row, and \(y_{<t}\) the prefix the student has already produced.
\Cref{eq:opsd-teacher} attaches replay-grounded guidance to the exact prefixes the current student has visited, so the supervision follows the student's own state distribution as it changes.
At each visited prefix, \OPSD minimizes the forward KL over vocabulary \(\mathcal{V}\)
\begin{equation}
\sum_{u\in\mathcal{V}}
q_{\bar\theta}^{e}(u\mid x,y_{<t})
\log
\frac{q_{\bar\theta}^{e}(u\mid x,y_{<t})}
{p_\theta(u\mid x,y_{<t})}.
\label{eq:opsd-forward-kl}
\end{equation}
where \(\mathcal{V}\) is the model vocabulary, \(q^{e}_{\bar\theta}\) the teacher distribution of \cref{eq:opsd-teacher}, and \(p_\theta\) the student distribution being optimized.
Training averages \cref{eq:opsd-forward-kl} over completions along each student-sampled response, masks prompt and padding tokens, and updates the student parameters \(\theta\).
The forward direction penalizes low probability on tokens the grounded teacher deems likely, keeping the student within the teacher’s support.

\paragraph{Rule-bot emulation.}
RuleBot-Distill provides a non-search control under the same student-facing interface.
It draws decision states and selected actions from trajectories generated by a fixed heuristic policy, yielding compact rule-verified completions.
Instead, \SPSD uses MuZero state and value predictors to construct a linearized MCTS trace containing the selected action, visit-derived preferences over alternatives, value estimates, and replayable branch evidence.
Both arms present the student with the same prompt format and require the same boxed legal handle. They differ in the teacher-side source of the action and its justification.
The comparison tests whether the search-derived structure improves on a verified action alone.
\Cref{app:prompts} and \cref{app:rulebot-distill-example,app:grounded-mcts-example} report the prompt templates and record examples for each configuration.
\Cref{app:config} details the search, optimization, decoding, and compute settings for all three paths.

\section{Experimental Setup}
\label{sec:experiments}

\paragraph{Training corpus and search experts.}
We construct the main \SPSD corpus from Connect4, Domineering, Simplified Othello, and Tic-Tac-Chess to cover alignment under gravity, spatial blocking, line capture, and movement-based tactics.
For each game, an EfficientZero expert~\citep{DBLP:conf/nips/YeLKAG21} is trained by self-play through LightZero~\citep{DBLP:conf/nips/NiuPYLZRHLL23}, with 50 MCTS simulations per decision during data generation.
Each game contributes 5,000 examples, reduced to 600 for Tic-Tac-Chess by its smaller pool of verified states.
Each game allocates 80\% of its examples to move choice with replay-grounded reasoning and 20\% to the six state-question tasks, which receive equal quotas after pooling.
Appendix~\ref{app:materialization} details the allocation and verification criteria, and Appendix~\ref{app:muzero-training} reports the expert-training settings.

\paragraph{Models and post-training.}
We test a pool of models spanning different families and types.
Qwen3-4B-Base, Qwen3-8B with thinking disabled and enabled, and Llama-3.1-8B cover base, thinking, and instruction-tuned conditions across two families.
SFT fits the linearized MCTS traces of the \SPSD corpus directly, and \OPSD supplies them as the privileged trace \(e\) that conditions the teacher.
RuleBot-Distill keeps the \OPSD procedure unchanged and replaces the search-derived corpus with completions from the heuristic rule bot, leaving the origin of the supervision as the only difference between the two conditions.
Every setting is trained for 1,000 optimization steps on one Nvidia H200 GPU with 141 GB.
Appendices~\ref{app:config} and~\ref{app:hyperparameters} detail the optimization, decoding, and compute settings.

\paragraph{Prompting.}
For move-choice tasks, all conditions receive the game rules, current state, player to move, and legal actions, and return a final action in the same boxed-answer format.
State-question prompts use the same state representation and require the task-specific answer in a box.
Appendices~\ref{app:prompts} and~\ref{app:spsd-dataset-taxonomy} specify the move prompt and state-question tasks.

\paragraph{Rule diversity.}
To test the effect of rule diversity under a fixed training budget, we compare curricula drawn from a separate pool of five base games and nine variants per game, yielding 50 environments.
Each variant changes a local rule governing scoring, openings, movement, or winning conditions while preserving the observation shape, board topology, and action-space size.
Appendix~\ref{app:variants} details the rule changes, and Appendix~\ref{app:stats} reports the composition and structural properties of these 50 environments.

\paragraph{Evaluation Benchmarks.}
We evaluate game transfer on 15 held-out games from the Ludii collection~\citep{DBLP:conf/ecai/PietteSSSWB20}, disjoint from the training environments.
Appendices~\ref{app:games} and~\ref{app:game-cards} describe the game selection and rules.
The primary game metric is the mean FIDE score, assigning 0, 0.5, and 1 to losses, draws, and wins, respectively.
Episode win rate and move legality are also reported, where legality is the fraction of scored attempts that produce a valid legal action.
Formatting, parsing, and illegal-action failures count as unsuccessful attempts.
For out-of-domain transfer, we evaluate mathematics on MATH500, AIME24, AIME25, AMC23, Olympiad Bench, and Minerva Math.
We use pass@1 for MATH500, Olympiad Bench, and Minerva Math, and average accuracy over 32 sampled responses per problem for AIME24, AIME25, and AMC23.
Appendix~\ref{app:config} specifies the sampling settings and generation limits for both evaluations.

\paragraph{Search distillation analysis.}
To assess whether \SPSD transfers the action-choice policy of MCTS into the LLM, which analyses candidate actions through the positions they lead to before committing to one, we follow~\citet{DBLP:journals/corr/abs-2605-06840} and use GPT-5.6 to read each reasoning trace and recover the positions it considers and the move it commits to, expressed as legal actions of the game.
Restoring the position the model faced at that turn and querying the expert of that game yields a value for every legal move together with the move its search prefers.
For each episode, \(Q_{50}\) averages the search values the expert assigns to the committed moves from the player-to-move perspective, over the turns for which it returns a value.
Oracle agreement measures how often the committed move matches the action the expert search prefers.
Appendix~\ref{app:q50-selection} reports the selection and aggregation protocol.

\begin{table*}[t]
\centering
\scriptsize
\setlength{\tabcolsep}{2.0pt}
\renewcommand{\arraystretch}{0.96}
\begin{threeparttable}
\resizebox{\textwidth}{!}{%
\begin{tabular}{@{}llcccccc|>{\columncolor{tableDeltaBg}}c|ccc@{}}
\toprule
\textbf{Model} & \textbf{Condition} & \textbf{MATH} & \textbf{A24} & \textbf{A25} & \textbf{AMC} & \textbf{Olympiad Bench} & \textbf{Minerva Math} & \textbf{$\Delta_\text{math}$} & \textbf{FIDE (\%)} & \textbf{Legal (\%)} & \textbf{Win (\%)} \\
\midrule
\multirow{4}{*}{\texttt{Qwen3-4B-Base}}
& \basecond & 56.4 & 6.2 & 5.9 & 29.7 & 26.4 & 20.2 & 24.1 & 15.0 & 65.0 & 15.0 \\
& \sftcond & \resultdelta{63.0}{+6.6} & \resultdelta{7.9}{+1.7} & \resultdelta{6.0}{+0.1} & \resultdelta{42.1}{+12.4} & \resultdelta{31.6}{+5.2} & \resultdelta{31.2}{+11.0} & \resultdelta{30.3}{+6.2} & \resultdelta{19.9}{+4.9} & \resultdelta{34.9}{-30.1} & \resultdelta{20.0}{+5.0} \\
& \rulebotcond & \secondbest{\resultdelta{69.4}{+13.0}} & \secondbest{\resultdelta{10.5}{+4.3}} & \secondbest{\resultdelta{9.6}{+3.7}} & \secondbest{\resultdelta{44.0}{+14.3}} & \best{\resultdelta{35.7}{+9.3}} & \secondbest{\resultdelta{34.2}{+14.0}} & \secondbest{\resultdelta{33.9}{+9.8}} & \secondbest{\resultdelta{35.1}{+20.1}} & \secondbest{\resultdelta{70.1}{+5.1}} & \secondbest{\resultdelta{35.0}{+20.0}} \\
& \opsdcond & \best{\resultdelta{74.0}{+17.6}} & \best{\resultdelta{12.7}{+6.5}} & \best{\resultdelta{12.4}{+6.5}} & \best{\resultdelta{47.0}{+17.3}} & \secondbest{\resultdelta{35.1}{+8.7}} & \best{\resultdelta{38.6}{+18.4}} & \best{\resultdelta{36.6}{+12.5}} & \best{\resultdelta{39.9}{+24.9}} & \best{\resultdelta{75.1}{+10.1}} & \best{\resultdelta{45.0}{+30.0}} \\
\midrule
\multirow{4}{*}{\texttt{Qwen3-8B}\,\thinkoff}
& \basecond & \secondbest{84.4} & 26.4 & 19.5 & 68.0 & 48.9 & 42.6 & 48.3 & 42.5 & \best{95.0} & 45.0 \\
& \sftcond & \resultdelta{83.2}{-1.2} & \resultdelta{24.4}{-2.0} & \resultdelta{19.9}{+0.4} & \resultdelta{64.2}{-3.8} & \resultdelta{48.1}{-0.8} & \resultdelta{41.2}{-1.4} & \resultdelta{46.8}{-1.5} & \best{\resultdelta{52.5}{+10.0}} & \best{\resultdelta{95.0}{0.0}} & \best{\resultdelta{55.0}{+10.0}} \\
& \rulebotcond & \secondbest{\resultdelta{84.4}{0.0}} & \best{\resultdelta{31.0}{+4.6}} & \secondbest{\resultdelta{21.4}{+1.9}} & \secondbest{\resultdelta{73.1}{+5.1}} & \best{\resultdelta{53.2}{+4.3}} & \best{\resultdelta{43.0}{+0.4}} & \secondbest{\resultdelta{51.0}{+2.7}} & \resultdelta{42.5}{0.0} & \resultdelta{90.0}{-5.0} & \resultdelta{45.0}{0.0} \\
& \opsdcond & \best{\resultdelta{86.8}{+2.4}} & \secondbest{\resultdelta{29.9}{+3.5}} & \best{\resultdelta{21.6}{+2.1}} & \best{\resultdelta{74.0}{+6.0}} & \secondbest{\resultdelta{52.9}{+4.0}} & \best{\resultdelta{43.0}{+0.4}} & \best{\resultdelta{51.4}{+3.1}} & \best{\resultdelta{52.5}{+10.0}} & \resultdelta{85.0}{-10.0} & \best{\resultdelta{55.0}{+10.0}} \\
\midrule
\multirow{4}{*}{\texttt{Qwen3-8B}\,\thinkon}
& \basecond & \secondbest{94.4} & \best{73.1} & 62.0 & \best{95.3} & 65.3 & \secondbest{59.2} & 74.9 & 51.0 & \best{100.0} & \projectedresult{53.7} \\
& \sftcond & \resultdelta{94.0}{-0.4} & \secondbest{\resultdelta{72.0}{-1.1}} & \resultdelta{64.0}{+2.0} & \secondbest{\resultdelta{94.8}{-0.5}} & \best{\resultdelta{67.7}{+2.4}} & \best{\resultdelta{59.9}{+0.7}} & \best{\resultdelta{75.4}{+0.5}} & \projectedresult{\resultdelta{15.6}{-35.4}} & \projectedresult{\resultdelta{30.6}{-69.4}} & \projectedresult{\resultdelta{16.4}{-37.3}} \\
& \rulebotcond & \secondbest{\resultdelta{94.4}{0.0}} & \resultdelta{70.3}{-2.8} & \secondbest{\resultdelta{64.2}{+2.2}} & \resultdelta{94.6}{-0.7} & \secondbest{\resultdelta{66.8}{+1.5}} & \resultdelta{58.8}{-0.4} & \resultdelta{74.9}{0.0} & \secondbest{\resultdelta{53.0}{+2.0}} & \best{\resultdelta{100.0}{0.0}} & \secondbest{\resultdelta{55.8}{+2.1}} \\
& \opsdcond & \best{\resultdelta{94.6}{+0.2}} & \resultdelta{70.9}{-2.2} & \best{\resultdelta{64.8}{+2.8}} & \resultdelta{94.1}{-1.2} & \secondbest{\resultdelta{66.8}{+1.5}} & \resultdelta{58.5}{-0.7} & \secondbest{\resultdelta{75.0}{+0.1}} & \best{\resultdelta{59.5}{+8.5}} & \best{\resultdelta{100.0}{0.0}} & \best{\resultdelta{62.7}{+9.0}} \\
\midrule
\multirow{4}{*}{\texttt{Llama-3.1-8B}}
& \basecond & 46.4 & \secondbest{4.3} & \best{1.5} & 21.7 & 15.7 & \best{27.6} & \secondbest{19.5} & 42.5 & 90.0 & 45.0 \\
& \sftcond & \best{\resultdelta{51.6}{+5.2}} & \resultdelta{2.6}{-1.7} & \resultdelta{0.6}{-0.9} & \resultdelta{23.4}{+1.7} & \resultdelta{16.0}{+0.3} & \secondbest{\resultdelta{27.2}{-0.4}} & \best{\resultdelta{20.2}{+0.7}} & \resultdelta{42.5}{0.0} & \best{\resultdelta{100.0}{+10.0}} & \resultdelta{45.0}{0.0} \\
& \rulebotcond & \secondbest{\resultdelta{47.0}{+0.6}} & \resultdelta{3.6}{-0.7} & \secondbest{\resultdelta{0.8}{-0.7}} & \secondbest{\resultdelta{23.9}{+2.2}} & \best{\resultdelta{17.6}{+1.9}} & \resultdelta{23.2}{-4.4} & \resultdelta{19.3}{-0.2} & \best{\resultdelta{47.5}{+5.0}} & \best{\resultdelta{100.0}{+10.0}} & \best{\resultdelta{50.0}{+5.0}} \\
& \opsdcond & \resultdelta{46.8}{+0.4} & \best{\resultdelta{4.9}{+0.6}} & \resultdelta{0.2}{-1.3} & \best{\resultdelta{24.9}{+3.2}} & \secondbest{\resultdelta{16.6}{+0.9}} & \resultdelta{21.7}{-5.9} & \resultdelta{19.2}{-0.3} & \best{\resultdelta{47.5}{+5.0}} & \best{\resultdelta{100.0}{+10.0}} & \best{\resultdelta{50.0}{+5.0}} \\
\bottomrule
\end{tabular}}
\begin{tablenotes}
\item[] \thinkon\,= thinking enabled; \thinkoff\,= thinking disabled.
\end{tablenotes}
\end{threeparttable}
\caption{\textbf{SPSD training effects.}
Mathematics benchmarks and their six-benchmark mean, normalized FIDE, legality, and win rate. \textbf{Best} and \secondbest{second to best} results are highlighted.}
\label{tab:main-results}
\end{table*}

\definecolor{trainingsft}{RGB}{48,98,170}
\definecolor{trainingrulebot}{RGB}{210,120,35}
\definecolor{trainingspsd}{RGB}{170,55,55}
\definecolor{trainingvariants}{RGB}{108,73,170}

\begin{figure*}[!t]
  \centering
  \resizebox{0.86\textwidth}{!}{%
  \begin{tikzpicture}
    \begin{axis}[
      name=fideaxis,
      width=0.255\textwidth,
      height=3.85cm,
      xmin=0,xmax=1000,
      ymin=10,ymax=45,
      xtick={0,250,500,750,1000},
      ytick={10,20,30,40},
      xlabel={\texttt{training step}},
      ylabel={\texttt{FIDE (\%)}},
      title={\texttt{(a) game performance}},
      title style={font=\fontsize{7.2}{7.2}\selectfont,yshift=-1ex},
      tick label style={font=\fontsize{6.1}{6.1}\selectfont},
      label style={font=\fontsize{6.8}{6.8}\selectfont},
      grid=major,
      major grid style={solid,gray!22},
      axis background/.style={fill=rqplotbg},
      legend to name=trainingprogresslegend,
      legend columns=4,
      legend style={draw=none,fill=none,font=\fontsize{5.6}{5.6}\selectfont,/tikz/every even column/.append style={column sep=0.8mm}},
    ]
      \addplot[trainingsft,line width=1.15pt,unbounded coords=discard,restrict expr to domain={\thisrow{MethodCode}}{1:1}]
        table[x=Step,y=FIDE,col sep=comma]{qwen3_4b_training_progress.csv};
      \addlegendentry{SFT}
      \addplot[trainingrulebot,line width=1.15pt,unbounded coords=discard,restrict expr to domain={\thisrow{MethodCode}}{2:2}]
        table[x=Step,y=FIDE,col sep=comma]{qwen3_4b_training_progress.csv};
      \addlegendentry{RuleBot-Distill}
      \addplot[trainingspsd,line width=1.15pt,unbounded coords=discard,restrict expr to domain={\thisrow{MethodCode}}{3:3}]
        table[x=Step,y=FIDE,col sep=comma]{qwen3_4b_training_progress.csv};
      \addlegendentry{OPSD (base)}
      \addplot[trainingvariants,line width=1.15pt,unbounded coords=discard,restrict expr to domain={\thisrow{MethodCode}}{4:4}]
        table[x=Step,y=FIDE,col sep=comma]{qwen3_4b_training_progress.csv};
      \addlegendentry{OPSD (+variants)}
    \end{axis}

    \begin{axis}[
      name=probeaxis,
      at={(fideaxis.east)},
      xshift=1.35cm,
      anchor=west,
      width=0.255\textwidth,
      height=3.85cm,
      xmin=0,xmax=1000,
      ymin=20,ymax=80,
      xtick={0,250,500,750,1000},
      ytick={20,40,60,80},
      xlabel={\texttt{training step}},
      ylabel={\texttt{legality rate (\%)}},
      title={\texttt{(b) move behavior}},
      title style={font=\fontsize{7.2}{7.2}\selectfont,yshift=-1ex},
      tick label style={font=\fontsize{6.1}{6.1}\selectfont},
      label style={font=\fontsize{6.8}{6.8}\selectfont},
      grid=major,
      major grid style={solid,gray!22},
      axis background/.style={fill=rqplotbg},
    ]
      \addplot[trainingsft,line width=1.15pt,unbounded coords=discard,restrict expr to domain={\thisrow{MethodCode}}{1:1},forget plot]
        table[x=Step,y=LegalityRate,col sep=comma]{qwen3_4b_training_progress.csv};
      \addplot[trainingrulebot,line width=1.15pt,unbounded coords=discard,restrict expr to domain={\thisrow{MethodCode}}{2:2},forget plot]
        table[x=Step,y=LegalityRate,col sep=comma]{qwen3_4b_training_progress.csv};
      \addplot[trainingspsd,line width=1.15pt,unbounded coords=discard,restrict expr to domain={\thisrow{MethodCode}}{3:3},forget plot]
        table[x=Step,y=LegalityRate,col sep=comma]{qwen3_4b_training_progress.csv};
      \addplot[trainingvariants,line width=1.15pt,unbounded coords=discard,restrict expr to domain={\thisrow{MethodCode}}{4:4},forget plot]
        table[x=Step,y=LegalityRate,col sep=comma]{qwen3_4b_training_progress.csv};
    \end{axis}

    \begin{axis}[
      name=mathaxis,
      at={(probeaxis.east)},
      xshift=1.35cm,
      anchor=west,
      width=0.255\textwidth,
      height=3.85cm,
      xmin=0,xmax=1000,
      ymin=20,ymax=40,
      xtick={0,250,500,750,1000},
      ytick={20,24,28,32,36,40},
      xlabel={\texttt{training step}},
      ylabel={\texttt{six-benchmark mean (\%)}},
      title={\texttt{(c) mathematics transfer}},
      title style={font=\fontsize{7.2}{7.2}\selectfont,yshift=-1ex},
      tick label style={font=\fontsize{6.1}{6.1}\selectfont},
      label style={font=\fontsize{6.8}{6.8}\selectfont},
      grid=major,
      major grid style={solid,gray!22},
      axis background/.style={fill=rqplotbg},
    ]
      \addplot[trainingsft,line width=1.15pt,unbounded coords=discard,restrict expr to domain={\thisrow{SeriesCode}}{1:1},forget plot]
        table[x=Step,y=MathAvg,col sep=comma]{qwen3_4b_training_progress.csv};
      \addplot[trainingspsd,line width=1.15pt,unbounded coords=discard,restrict expr to domain={\thisrow{SeriesCode}}{2:2},forget plot]
        table[x=Step,y=MathAvg,col sep=comma]{qwen3_4b_training_progress.csv};
      \addplot[trainingrulebot,line width=1.15pt,unbounded coords=discard,restrict expr to domain={\thisrow{SeriesCode}}{3:3},forget plot]
        table[x=Step,y=MathAvg,col sep=comma]{qwen3_4b_training_progress.csv};
      \addplot[trainingvariants,line width=1.15pt,unbounded coords=discard,restrict expr to domain={\thisrow{SeriesCode}}{4:4},forget plot]
        table[x=Step,y=MathAvg,col sep=comma]{qwen3_4b_training_progress.csv};
    \end{axis}
  \end{tikzpicture}%
  }

  \pgfplotslegendfromname{trainingprogresslegend}
  \caption{\textbf{Training trajectories.} Evolution of reasoning abilities of Qwen3-4B-Base during training. \SPSD's advantage emerges from opposite dynamics: SFT peaks early and declines while OPSD keeps improving.}
  \label{fig:qwen3-4b-training-progress}
\end{figure*}
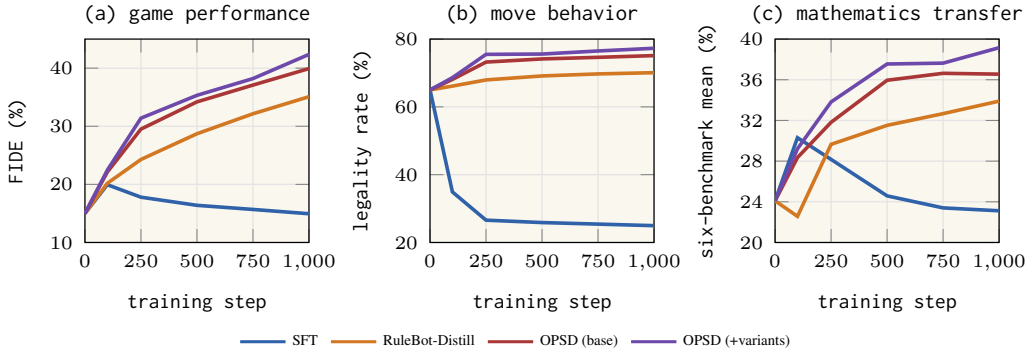

\section{Results}
\label{sec:results}
\label{sec:metrics-analysis}

Our results support a conditional view of search transfer.
Search-derived supervision improves performance when the student can convert value-ordered traces into decisions, and neither legality nor longer reasoning traces is sufficient on its own.
Table~\ref{tab:main-results}, Figure~\ref{fig:qwen3-4b-training-progress}, and Figure~\ref{fig:rq5-geometry} expose this distinction across model conditions, throughout optimization, and under fresh search replay.

\paragraph{\rq1 How does transfer vary across model families and decision metrics?}
Held-out games provide the central test of transfer beyond the training environments, where the model must choose actions in unfamiliar rule systems (Table~\ref{tab:main-results}).
For Qwen3-4B-Base, \OPSD triples the held-out-game win rate from 15\% to 45\%.
The corresponding FIDE score rises from 15\% to 39.9\%, while the model produces legal actions in 75.1\% of the evaluated positions.
Search-distilled guidance lets the model use ranked alternatives and their consequences when it encounters games outside the training environments.
This knowledge transfers to mathematical reasoning, raising the six-benchmark mean from 24.1 to 36.6.
The two evaluation gains mark a reusable reasoning pattern learned from synthetic search supervision.
The Qwen3-8B conditions show how thinking configuration shapes this transfer.
With thinking enabled, mathematics performance is already near saturation in this comparison, while \OPSD achieves the strongest trained held-out-game result with complete action legality.
With thinking disabled, search-derived knowledge transfers to both held-out games and mathematical reasoning.
The instruction-tuned Llama-3.1-8B condition produces the same game outcomes under both expert sources.

\begingroup
\pgfplotsset{
  rqfiveaxis/.style={
    width=\linewidth,
    height=3.35cm,
    xmin=-1.0,xmax=1.1,
    ymin=0,ymax=1.0,
    zmin=0,zmax=1.03,
    xtick={-1,-0.5,0,0.5,1},
    ytick={0,0.5,1},
    ztick={0,0.5,1},
    xlabel={$Q_{50}$},
    ylabel={\texttt{oracle@50}},
    zlabel={\texttt{smoothed win}},
    tick label style={font=\fontsize{6.0}{6.0}\selectfont},
    label style={font=\fontsize{6.2}{6.2}\selectfont},
    grid=major,
    major grid style={dashed,gray!42},
    axis background/.style={fill=rqplotbg},
    view={50}{32},
    colormap name=viridis,
    clip=false,
  },
}

\newcommand{\rqfivepanel}[2]{%
  \begin{subfigure}[t]{0.245\textwidth}
    \centering
    \begin{tikzpicture}
      \begin{axis}[rqfiveaxis]
        \addplot3[
          surf,
          mesh/rows=30,
          mesh/cols=30,
          mesh/ordering=x varies,
          shader=interp,
          opacity=0.72,
          draw=black!18,
          point meta=explicit,
          forget plot
        ] table[
          x=Q50,
          y=OracleAgreement50,
          z=PlotWinRate,
          meta=PlotWinRate,
          col sep=comma
        ] {q50_surface_grid_#1.csv};
      \end{axis}
    \end{tikzpicture}
    \captionsetup{font=footnotesize,margin=0mm}
    \subcaption{\texttt{#2}}
  \end{subfigure}%
}

\begin{figure*}[!t]
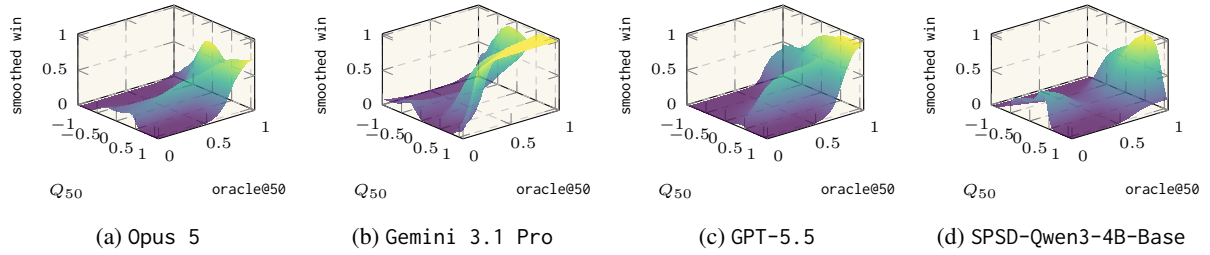

  \centering
  \rqfivepanel{opus}{Opus 5}
  \hfill
  \rqfivepanel{gemini}{Gemini 3.1 Pro}
  \hfill
  \rqfivepanel{gpt55}{GPT-5.5}
  \hfill
  \rqfivepanel{spsd}{SPSD-Qwen3-4B-Base}
  \caption{\textbf{MCTS structure and reasoning sharpness.} Surfaces relate episode wins to mean selected-action value $Q_{50}$ and \texttt{oracle@50} under a fresh 50-simulation MuZero search.\vspace{0.7cm}}
  \label{fig:rq5-geometry}
\vspace{-\baselineskip}
\end{figure*}
\endgroup

\paragraph{\rq2 How does the expert source affect transfer?}
For Qwen3-4B-Base, \OPSD raises held-out-game win rate from 35\% with RuleBot-Distill to 45\%.
The same condition raises the six-benchmark mathematics mean from 33.9 to 36.6 (Table~\ref{tab:main-results}).
\OPSD is the strongest overall solution among the compared Qwen3-4B training conditions, with the highest mathematics aggregate, FIDE, legality, and win rate.
RuleBot-Distill achieves the best OlympiadBench result among these conditions, scoring 35.7 compared with 35.1 for \OPSD.
The accompanying mathematics improvement extends the gain to problems where comparing candidate continuations and rejecting unproductive ones can guide a solution.

\paragraph{\rq3 How does transfer evolve during training?}
To trace how game behavior and mathematics evolve during training, we compare the best observed performance at each training budget in Figure~\ref{fig:qwen3-4b-training-progress}.
SFT learns the linearized targets quickly, but its held-out-game performance peaks early and declines as training continues.
The loss of earlier gains is consistent with increasing forgetting under longer continual SFT in this setting.
The concurrent decline in legality and mathematics suggests that continued SFT narrows the model's useful behavior across both the game interface and the separate reasoning tasks.
\OPSD improves later in training by attaching search evidence to prefixes sampled by the current model, keeping the update target connected to the decisions the model must make.
At step 1000, the variants curriculum reaches 42.4\% FIDE, 77.3\% legality, and a mathematics mean of 39.2, compared with 39.9\%, 75.1\%, and 36.6 for the base curriculum.
The variants condition leads across all three metrics at every evaluated post-training step, supporting broader coverage of search-grounded decision patterns.
Its advantage on both evaluations suggests that variation in the training rules encourages decision patterns that remain useful across environments and reasoning tasks.
This result motivates using \SPSD to diversify the decisions and consequences encountered during post-training while preserving guidance on the model's own responses.

\paragraph{\rq4 How does \SPSD distill MCTS search into model decisions?}
We investigate how \SPSD distills MCTS search by relating the model's selected actions to MCTS values and preferences, and examining their association with game outcomes (Figure~\ref{fig:rq5-geometry}).
Wins concentrate in the high-$Q_{50}$ and high-agreement region across the four systems.
At high selected-action values, the smoothed win surfaces rise with agreement, linking successful play to the model's preference for the actions favored by MCTS.
Gemini 3.1 Pro maintains a high estimated win rate at intermediate agreement when its selected actions have high search values, indicating that successful choices can differ from those preferred by MCTS.
For GPT-5.5 and \SPSD, high-valued choices are more strongly associated with wins when agreement is also high, while Opus 5 has a lower estimated win rate even when both value and agreement are high.
For the \SPSD model, successful play is associated with generating the actions favored by MCTS.
The stronger coupling between successful play and MCTS agreement than in Gemini 3.1 Pro suggests that our comparative search supervision encourages a reasoning pattern centered on ranking actions by their downstream consequences.
The model makes these decisions from the game prompt while the search values and traces remain hidden at inference, so agreement with the independently evaluated search reflects alignment in the choices it produces.
This alignment provides a behavioral signature of distillation, with the model favoring alternatives whose evaluated consequences support successful play.

\section{Conclusion}
\label{sec:conclusion}

\SPSD turns records produced by self-play search experts in executable environments into replay-grounded supervision for language-model post-training.
Each record combines the selected action with ranked alternatives, search values, and replayed continuations to support comparative move-choice traces, verified state questions, and \OPSD along student-generated prefixes.
Across the Qwen3-4B training budgets studied, \OPSD provides the most stable overall performance improvements among the compared methods, while SFT loses earlier gains under longer training.
The variant-enriched curriculum strengthens held-out-game performance and mathematical reasoning transfer through more diverse search-grounded supervision.
\SPSD opens a direction for future work on continual training through expanding curricula of executable environments, with search experts generating supervision for new rules and decision problems.

\section*{Acknowledgements}
This research was partially supported by the AI-PACT project (CUP B47H22004450008, B47H22004460001); the National Plan PNC-I.1 DARE initiative (PNC0000002, CUP B53C22006450001); the Horizon Europe VALWARP project (proposal no. 101326664); Huawei–Edinburgh Joint Research Laboratory Grant. We thank LG Solution Srl for partially funding a PhD scholarship to L. Molfetta.

\bibliographystyle{acl_natbib}
\bibliography{custom}

\newpage
\clearpage
\appendix
\raggedbottom
\onecolumn
\setcounter{topnumber}{4}
\setcounter{bottomnumber}{2}
\setcounter{totalnumber}{5}
\renewcommand{\topfraction}{0.95}
\renewcommand{\bottomfraction}{0.95}
\renewcommand{\textfraction}{0.05}
\renewcommand{\floatpagefraction}{0.6}
\section{Grounded MCTS Linearization}
\label{app:grounded-linearization}

To distill search reasoning abilities into the LLM, we linearize the records that the trained \SPSD MuZero experts produce.
For each decision state, a non-terminal board position with the player to move and the legal actions available in it, the experts export one record.
We state the recorded decision, its retained alternatives, and their consequences in natural language.
Replay restores the recorded state and follows each retained branch, so every statement in the resulting trace follows from an observed transition.
A record \(m(s)\) stores the serialized environment state \(s\), its legal-action set, the recorded target, the root policy and value outputs, and, for each exported root action, its canonical handle, visit count, export index, action-specific edge value or missing-value marker, and retained branch prefix.
Numerical search telemetry selects which evidence to retain and never enters the trace.
The record pairs the trace with the action sequence it replays, and only the trace reaches the student, ending with exactly one boxed legal handle.

\paragraph{Replay validation.}
Restoring \(s\) in the executable environment recomputes the board, active player, legal actions, and phase-specific fields.
A record is retained only when these fields agree with their serialized values and \(a^\star\in\mathcal{A}(s)\).
Failure to replay \(a^\star\) rejects the record, whereas failure to replay an optional alternative or reply removes only that branch from the trace.
For every retained branch, replay records the legal action prefix, the recomputed successor states, and any terminal flag and outcome reached along the prefix.

\paragraph{Branch selection and evidence gates.}
Each export contains at most eight root-action records.
After excluding \(a^\star\), the exported legal action with the greatest visit count becomes \(a^{\mathrm{alt}}\), with ties broken by export order.
The corresponding edge value is
\[
Q(a)=r(s,a)+\gamma\sigma V(T(s,a)),
\]
expressed from the current root player's perspective, where \(\gamma=1\) in this corpus, \(\sigma=-1\) when the child changes player, and \(\sigma=+1\) otherwise.
Values are not rescaled across states or games, and a missing exported value is represented by \(\bot\).
If the exported subset contains no other legal action, it sets \(a^{\mathrm{alt}}=\bot\) and \(Q(\bot)=\bot\) and omits the contrast.
Target-only detail requires either a numeric \(Q(a^\star)\) or a replayed terminal win.
A target--alternative contrast additionally requires a retained alternative and either \(Q(a^\star)-Q(a^{\mathrm{alt}})\geq\Delta\), with \(\Delta=0.05\), or a better replayed terminal outcome for the target.
Terminal outcomes are ordered win \(>\) draw \(>\) loss from the root player's perspective.

\begin{algorithm*}[t]
\caption{\textbf{Grounded MCTS linearization.} Replay validates the selected action, any retained alternative, and any bounded reply before emitting a fallback, target-only, or contrast trace ending in exactly one boxed legal handle.}
\label{alg:grounded-mcts-linearization}
\begin{algorithmic}[1]
\Require Search record \(m(s)\) with serialized state \(s\), legal set \(\mathcal{A}(s)\), target \(a^\star\), up to eight exported root-action records, replay depth \(D=1\), separation threshold \(\Delta=0.05\)
\State Restore \(s\) and recompute its state fields and \(\mathcal{A}(s)\); reject if any field disagrees with the serialized record or \(a^\star\notin\mathcal{A}(s)\).
\State Replay \(a^\star\); reject on failure, otherwise record its state effect and any observed terminal outcome.
\If{an exported legal alternative exists}
\State Choose \(a^{\mathrm{alt}}\neq a^\star\) with greatest visit count, preserving export order for ties.
\State Replay its retained branch prefix from \(s\); if any action is illegal, set \(a^{\mathrm{alt}}\gets\bot\) and \(Q(a^{\mathrm{alt}})\gets\bot\), otherwise record the recomputed states and any terminal outcome.
\Else
\State Set \(a^{\mathrm{alt}}\gets\bot\) and \(Q(a^{\mathrm{alt}})\gets\bot\).
\EndIf
\State Set the detail gate \(d\gets[Q(a^\star)\text{ exists}]\lor[\text{the target replays to a win}]\).
\State Set the contrast gate \(c\gets[a^{\mathrm{alt}}\neq\bot]\land([Q(a^\star)-Q(a^{\mathrm{alt}})\geq\Delta]\lor[\text{the target has a better replayed terminal outcome}])\), with the numeric test false when either value is missing.
\State Retain target facts from replay; select contrast mode if \(c\), target-only mode if \(\neg c\land d\), and fallback mode otherwise.
\If{the target is non-terminal and \(d\) is true}
\State Follow at most \(D\) replayed opponent replies on the target branch.
\State Drop the reply walk if any action is illegal; otherwise retain only facts recomputed from the replayed states.
\EndIf
\State Attach win, loss, or draw language only when replay reaches the corresponding terminal outcome.
\State Write the selected mode from the recomputed facts, including any retained reply and the commitment to \(a^\star\).
\State Append exactly one final line containing \(\boxed{a^\star}\).
\State \Return the trace and replayed action sequence.
\end{algorithmic}
\end{algorithm*}

\paragraph{Trace modes.}
Fallback mode states only replay-supported target facts and the required final action.
Target-only mode adds detail about the selected action when the detail gate is satisfied, while contrast mode compares it with \(a^{\mathrm{alt}}\) only when the contrast gate is satisfied.
Illegal replies are omitted, and non-terminal prefixes are described without outcome language.
Retained facts take a stable surface form across states and games.

\section{Corpus Materialization}
\label{app:materialization}

To build supervised rows from the exported records, materialization produces one move-choice family and six state-QA families.
Each row is retained only when it passes the executable checks in Table~\ref{tab:row-verification}.

\begin{table*}[t]
\centering
\scriptsize
\setlength{\tabcolsep}{4.2pt}
\begin{tabularx}{\textwidth}{@{}l l X@{}}
\toprule
\textbf{Row family} & \textbf{Reference output} & \textbf{Executable criterion} \\
\midrule
Move choice & One boxed legal handle & The recomputed legal set contains the target; every narrated branch replays legally; any outcome statement follows an observed terminal state \\
Occupancy & Piece or empty-cell identity & Board value at the queried coordinate after state restoration \\
Legality & Boolean action status & Membership in the legal-action set recomputed from the restored state \\
Threat count & Integer count & Environment-specific winning threats enumerated from legal successors \\
Legal-action count & Integer count & Cardinality of the recomputed legal-action set \\
Legal-action enumeration & Complete handle set & Exact set equality with the legal-action handles enumerated from the restored state \\
Successor state & Next-state board text & One legal action replayed through the environment transition function \\
\bottomrule
\end{tabularx}
\caption{\textbf{Executable criteria for supervision rows.} Each reference output is checked against a restored state or replayed successor. Move-choice rows additionally require legal replay of every narrated branch and terminal evidence for every outcome claim.}
\label{tab:row-verification}
\end{table*}

\paragraph{Record interface and source identity.}
Each stored row is written as \((x,y;e)\), where \(x\) contains the state and task prompt, \(y\) the reference response, and \(e\) teacher-side reasoning or verifier context hidden from the student input.
The selected move, reference answer, search telemetry, and replay context remain in \(y\) or \(e\), never in \(x\).
The source-state identifier is a SHA-256 digest of a deterministic family-specific representation containing the board, active player, legal actions, and the rule or phase metadata required to reconstruct the state.
The same identifier is carried by the move-choice and state-QA families.

\paragraph{Allocation and trajectory generation.}
Materialization first reserves an equal number of rows for each of the six state tasks, then draws move-choice rows from the remaining decision states.
Each game contributes 100 trajectories, each opening with a non-terminal prefix of at most eight moves drawn uniformly at random from the legal actions of each state, after which the expert controls both seats.
Prefix actions provide state diversity and are excluded from the training rows.

\paragraph{Canonicalization and provenance.}
After every environment operation, legal handles and answer formats are canonicalized.
We then order candidate states by a seed-keyed hash of the source identity, move index, and decision-state identity, and assign them to task families and split groups in that order. Move-choice rows are drawn last.
RuleBot-Distill draws from rule-bot trajectories, whereas \SPSD draws from randomized-prefix self-play.
The two sources share the same row interface and keep distinct teacher and decision-state provenance.

\section{Experimental Configuration}
\label{app:config}

Tables~\ref{tab:expert-data-config}--\ref{tab:decode-compute-config} report the settings for expert-data generation, language-model post-training, benchmark decoding, and hardware allocation.
Replay geometry is evaluated with 50-simulation \(Q_{50}\) estimates.

\begin{table*}[t]
\centering
\scriptsize
\setlength{\tabcolsep}{5pt}
\begin{tabularx}{\textwidth}{@{}l l X@{}}
\toprule
\textbf{Component} & \textbf{Setting} & \textbf{Selected value} \\
\midrule
Expert & Family / search procedure & EfficientZero-trained MuZero experts / full-tree search \\
Expert search & Simulations per retained move & \(50\) \\
State sampling & Opening diversification & Uniformly sampled legal prefix of at most 8 moves (seed 0); prefix actions excluded from training rows \\
State sampling & Expert control & The expert controls both seats after the prefix \\
Move choice & Exported roots / retained alternatives & At most 8 / at most 1 \\
Move choice & Reply depth & \(D=1\) \\
Move choice & Root separation & \(\Delta=0.05\) \\
Move choice & Surface content & Legal targets; retained alternatives and replies; replay-observed terminal outcomes \\
Corpus & Source games & Connect4, Domineering, Simplified Othello, Tic-Tac-Chess \\
Corpus & Row-family separation & Move-choice and state-QA rows use disjoint unique decision states \\
State tasks & Balance rule & Equal quotas for occupancy, legality, threat count, legal-action count, legal-action enumeration, and successor state \\
\bottomrule
\end{tabularx}
\caption{\textbf{Expert-data configuration.} Search, state-sampling, move-choice, and allocation settings used to materialize the \SPSD corpus.}
\label{tab:expert-data-config}
\end{table*}

\begin{table*}[t]
\centering
\scriptsize
\setlength{\tabcolsep}{4.5pt}
\begin{tabularx}{\textwidth}{@{}l l X@{}}
\toprule
\textbf{Stage} & \textbf{Setting} & \textbf{Selected value} \\
\midrule
Reasoning-chain SFT & Sequence length / learning rate & 4,096 tokens / \(2\times10^{-5}\) \\
Reasoning-chain SFT & Epochs / effective batch & 2 / 128 examples \\
\midrule
\OPSD & Base learning rate / schedule & \(5\times10^{-6}\) / constant \\
\OPSD & Distillation objective / teacher update & Full-vocabulary forward KL / fixed teacher \\
\OPSD & Optimizer / steps / effective batch & AdamW / 1,000 / 16 \\
\OPSD & Completion / sequence limit & 1,024 tokens / 20,000 tokens \\
\OPSD & Gradient norm / divergence clip & 0.1 / 0.06 \\
\OPSD & Distillation weight / rollout sampling & \(\lambda=1.0\) / \(T=1.1,\ p=0.95,\ k=20\) \\
\OPSD & LoRA adapter & \(r=64,\ \alpha=128,\) dropout \(=0\) \\
\OPSD & Student / teacher thinking & Disabled / enabled \\
\OPSD & Seed & 42 \\
\midrule
Game benchmark & Prompt / answer contract & Rules, state, side to move, and legal handles; exact boxed-answer format; thinking disabled by default and enabled for the native-thinking row \\
Game benchmark & Sampling / limits / failures & \(T=0.7,\ p=0.9\); 16,384-token output and 24,576-token model limits; request retry limit 2; invalid or truncated actions score zero \\
Mathematics & Suites & Pass@1: MATH500, OlympiadBench, Minerva Math; AVG@32: AIME24, AIME25, AMC23 \\
Mathematics & Decoding / stopping & Boxed-answer decoding with constrained continuation; \(T=0.6,\ p=0.95\); 8,192-token limit; seed 0; tokenizer EOS and template terminators; unresolved format failures remain incorrect \\
Compute & Hardware & Expert training: 1 H200 GPU \\
\bottomrule
\end{tabularx}
\caption{\textbf{Post-training, decoding, and compute configuration.} Selected reasoning-chain SFT and \OPSD hyperparameters, fixed game and mathematics decoding contracts, and hardware allocation.}
\label{tab:decode-compute-config}
\end{table*}

\subsection{MuZero-family Expert Training}
\label{app:muzero-training}

The language-data teachers are EfficientZero agents, a MuZero-family algorithm, trained by self-play.
Table~\ref{tab:muzero-training} reports the settings that determine the four experts.

\begin{table*}[t]
\centering
\scriptsize
\setlength{\tabcolsep}{3.5pt}
\begin{tabularx}{\textwidth}{@{}l l X@{}}
\toprule
\textbf{Game} & \textbf{Setting} & \textbf{Value} \\
\midrule
All four & Algorithm / mode & EfficientZero / self-play for both seats \\
All four & Search / batch / updates & 50 simulations per move / batch 256 / 50 learner updates per collection \\
All four & Optimizer / learning rate / seed & Adam / $0.003$ constant learning rate / 0 \\
All four & Unroll / reanalyse / replay & 5 steps / ratio 0 / replay capacity 100,000 segments \\
All four & Model & One residual block, 64 channels \\
All four & Environment workers & 8 collectors, 5 evaluators, 5 evaluation episodes \\
All four & Evaluation & Rule-bot action interface; evaluation every 2,000 training iterations \\
Connect4 & Environment steps / segment geometry & 500,000 / LightZero default segment and TD geometry \\
Domineering & Environment steps / scheduler & 500,000 / piecewise-decay scheduler disabled \\
Simplified Othello & Environment steps / segment geometry / collectors & 1,000,000 / segment length 64 and TD horizon 64 / 32 collectors \\
Tic-Tac-Chess & Environment steps / scheduler & 500,000 / piecewise-decay scheduler disabled \\
\bottomrule
\end{tabularx}
\caption{\textbf{EfficientZero expert-training configuration.} The four teachers share the common profile unless a game-specific override is shown. The retained data comes from a separate 50-simulation search run from each expert.}
\label{tab:muzero-training}
\end{table*}
The training runs use undiscounted board-game returns ($\gamma=1$), deterministic argmax evaluation, and no reanalysis. Root noise and temperature affect collection only through the LightZero defaults. Data generation then applies the trained expert's deterministic 50-simulation policy after the randomized opening prefix.

\subsection{Replay-Geometry Expert Selection}
\label{app:q50-selection}

We evaluate the four experts at training steps 50, 100, 200, and 400 on a shared set of scenario indices.
Game-macro oracle agreement selects the expert, with game-macro \(Q_{50}\) breaking ties.
The selected expert defines the replay surface used in Figure~\ref{fig:rq5-geometry}.
Scoreable turns are averaged by episode and episodes with missing \(Q_{50}\) are omitted.

\section{Controlled Variants}
\label{app:variants}

The controlled-diversity intervention compares two curricula under the same 1,000-step optimization budget.
Each rule family contributes 9 variants that alter local rules while preserving the observation shape, board topology, and action space size.
Controlled variants mutate one local factor at a time along the axes listed in Table~\ref{tab:variant-axes}, so that transfer can be attributed to a specific perturbation.
This rule-family roster is reported separately from the four-game main corpus in Table~\ref{tab:expert-data-config}.

\begin{table}[!h]
\centering
\scriptsize
\setlength{\tabcolsep}{3.5pt}
\begin{tabularx}{\linewidth}{@{}l X@{}}
\toprule
\textbf{Parent family} & \textbf{Controlled rule axes} \\
\midrule
3~6~9 & Scoring thresholds and weights; orthogonal versus diagonal scoring; center or corner locks \\
Connect4 & Exact versus gapped four; legal winning directions; opening restrictions; locked columns \\
Domineering & Player orientations; shared modes; contact and opening constraints; blocked-state outcome \\
First Attack & Forbidden-neighbor geometry; opening locks; normal versus misere last move; move offsets \\
Tic-Tac-Chess & Placement timing; winning-line direction; piece movement; hop or capture behavior; locks \\
\bottomrule
\end{tabularx}
\caption{\textbf{Controlled rule axes.} Each row lists the local rule factors varied within one family. Every admitted variant preserves the parent observation shape and legal-handle space.}
\label{tab:variant-axes}
\end{table}

The model architecture, optimizer, training schedule, prompt contract, and evaluation protocol remain fixed across both curricula.
A variant is admitted only when it preserves the parent observation shape and legal-handle space and passes the executable-state checks.

\section{Game Corpus}
\label{app:games}

The controlled-diversity corpus comprises five base games and 15 held-out games, and the held-out set supplies the main-text evaluation.
We selected the games from the Ludii collection~\citep{DBLP:conf/ecai/PietteSSSWB20}, keeping finite two-player games with compact board states, discrete legal actions, and terminal outcomes that the \SPSD game interface can write and check.
Each selected game became a language-facing environment with a normalized board form, legal-action enumerator, transition function, terminal predicate, and outcome function.
GPT-5.6-sol and Opus 4.8 generated the rule descriptions and the environment code in cooperation, and each model's output was cross-validated against the other's before the game entered the corpus.
For each game the generated environment normalizes the rule text, defines the prompt-facing state notation, enumerates legal actions, implements deterministic transition and terminal checks, and verifies that its rollouts match the intended rules.
A candidate enters the library only after it passes schema validation, legality checks, invariant rollouts, and smoke training.
The five \SPSD base games are Connect4, Tic-Tac-Chess, 3~6~9, Domineering, and First Attack.
\subsection{Game Cards}
\label{app:game-cards}

Each card shows the prompt-facing rules of one game and a representative board sketch.
\newcommand{\gctagsThreeSixNine}{\gcmeta{shared placement}{higher score}}
\newcommand{\gctagsSimpleGame}{\gcmeta{orthogonal steps}{line of three}}
\newcommand{\gctagsAgapi}{\gcmeta{bishop move + shot}{immobilize}}
\newcommand{\gctagsAllQueens}{\gcmeta{queen slides}{line of three}}
\newcommand{\gctagsAlquerkonane}{\gcmeta{diagonal moves/jumps}{immobilize}}
\newcommand{\gctagsAralzaa}{\gcmeta{forward leaps}{opponent row}}
\newcommand{\gctagsBajr}{\gcmeta{forward steps}{opposite camp}}
\newcommand{\gctagsChains}{\gcmeta{place/merge/capture}{largest chain}}
\newcommand{\gctagsConnectFour}{\gcmeta{column drop}{connect four}}
\newcommand{\gctagsDomineering}{\gcmeta{domino placement}{last move}}
\newcommand{\gctagsEpelle}{\gcmeta{adjacent slides}{new line of three}}
\newcommand{\gctagsFeldja}{\gcmeta{place/slide/remove}{capture}}
\newcommand{\gctagsFirstAttack}{\gcmeta{non-attacking place}{last move}}
\newcommand{\gctagsGroups}{\gcmeta{move/jump}{connected group}}
\newcommand{\gctagsKingsValley}{\gcmeta{maximal slides}{king to throne}}
\newcommand{\gctagsMoxie}{\gcmeta{place/move/jump}{line or capture}}
\newcommand{\gctagsOthello}{\gcmeta{outflank flips}{disc majority}}
\newcommand{\gctagsTapatan}{\gcmeta{place/slide}{line of three}}
\newcommand{\gctagsTicTacChess}{\gcmeta{place/queen moves}{line of three}}
\newcommand{\gctagsTicTacToe}{\gcmeta{mark placement}{line of three}}

\newcommand{\gcgrid}[1]{%
  \fill[cardpaper] (0,0) rectangle (#1,#1);
  \draw[black!18, line width=0.25pt, step=1] (0,0) grid (#1,#1);
  \draw[cardink, line width=0.45pt] (0,0) rectangle (#1,#1);
}
\newcommand{\gcstoneone}[2]{\fill[cardp1] (#1,#2) circle (0.24);}
\newcommand{\gcstonetwo}[2]{\fill[cardp2] (#1,#2) circle (0.24);}
\newcommand{\gcemptypt}[2]{\draw[fill=cardpaper, draw=black!30, line width=0.35pt] (#1,#2) circle (0.10);}
\newcommand{\gcstoneptone}[2]{\draw[fill=cardp1, draw=cardink, line width=0.25pt] (#1,#2) circle (0.16);}
\newcommand{\gcstonepttwo}[2]{\draw[fill=cardp2, draw=cardink, line width=0.25pt] (#1,#2) circle (0.16);}
\newcommand{\gcblock}[2]{\fill[black!25] (#1,#2) rectangle ++(1,1);}
\newcommand{\gcchessqueen}[3]{\node[font=\fontsize{16}{16}\selectfont,text=#3] at (#1,#2) {\symqueen};}
\newcommand{\gcchessking}[3]{\node[font=\fontsize{16}{16}\selectfont,text=#3] at (#1,#2) {\symking};}
\newcommand{\gcchesspawn}[3]{\node[font=\fontsize{15}{15}\selectfont,text=#3] at (#1,#2) {\sympawn};}
\newcommand{\gcchessknight}[3]{\node[font=\fontsize{15}{15}\selectfont,text=#3] at (#1,#2) {\symknight};}
\newcommand{\gchexcell}[3]{%
  \filldraw[fill=#3, draw=black!25, line width=0.35pt]
    ({#1+0.42},{#2}) --
    ({#1+0.21},{#2+0.36}) --
    ({#1-0.21},{#2+0.36}) --
    ({#1-0.42},{#2}) --
    ({#1-0.21},{#2-0.36}) --
    ({#1+0.21},{#2-0.36}) -- cycle;}

\newcommand{\gameboardFeldja}{%
  \begin{tikzpicture}[scale=0.62]
    \draw[black!45, line width=0.8pt] (0,0) rectangle (4,4);
    \draw[black!45, line width=0.8pt] (0.7,0.7) rectangle (3.3,3.3);
    \draw[black!45, line width=0.8pt] (1.4,1.4) rectangle (2.6,2.6);
    \draw[black!45, line width=0.8pt] (2,4)--(2,3.3) (4,2)--(3.3,2) (2,0)--(2,0.7) (0,2)--(0.7,2);
    \foreach \x/\y in {0/0,2/0,4/0,0/2,4/2,0/4,2/4,4/4,0.7/0.7,2/0.7,3.3/0.7,0.7/2,3.3/2,0.7/3.3,2/3.3,3.3/3.3,1.4/1.4,2/1.4,2.6/1.4,1.4/2,2.6/2,1.4/2.6,2/2.6,2.6/2.6}{\gcemptypt{\x}{\y}}
    \gcstoneptone{0}{4}\gcstoneptone{2}{4}\gcstoneptone{4}{4}\gcstoneptone{2}{0.7}
    \gcstonepttwo{0.7}{3.3}\gcstonepttwo{3.3}{3.3}\gcstonepttwo{2.6}{1.4}
    \draw[cardp1, line width=1.0pt] (0,4)--(2,4)--(4,4);
  \end{tikzpicture}%
}

\newcommand{\gameboardMoxie}{%
  \begin{tikzpicture}[scale=0.78]
    \gcgrid{3}
    \fill[cardp2!10] (1,1) rectangle (2,2);
    \fill[cardp1!12] (2,1) rectangle (3,2);
    \draw[black!18, line width=0.25pt, step=1] (0,0) grid (3,3);
    \draw[cardink, line width=0.45pt] (0,0) rectangle (3,3);
    \gcchesspawn{0.5}{2.5}{cardp1}
    \gcchesspawn{0.5}{1.5}{cardp1}
    \gcchesspawn{2.5}{0.5}{cardp1}
    \gcchesspawn{2.5}{2.5}{cardp2}
    \gcchesspawn{1.5}{1.5}{cardp2}
    \draw[cardp2!75, line width=0.65pt] (1.5,1.5) circle (0.29);
    \draw[cardp1, line width=0.9pt, -latex] (0.5,1.5)--(2.5,1.5);
  \end{tikzpicture}%
}

\newcommand{\gameboardAllQueens}{%
  \begin{tikzpicture}[scale=0.68]
    \gcgrid{4}
    \foreach \x/\y in {1/3,2/2,0/1,1/0}{\gcblock{\x}{\y}}
    \foreach \x/\y in {0.5/3.5,1.5/1.5,2.5/1.5}{\gcchessqueen{\x}{\y}{cardp1}}
    \foreach \x/\y in {3.5/3.5,0.5/0.5,2.5/0.5}{\gcchessqueen{\x}{\y}{cardp2}}
    \draw[cardp1, line width=0.8pt, -latex] (1.5,3.5)--(1.5,1.85);
  \end{tikzpicture}%
}

\newcommand{\gameboardBajr}{%
  \begin{tikzpicture}[scale=0.68]
    \gcgrid{4}
    \fill[cardp2!10] (0,0) rectangle (2,1);
    \fill[cardp2!10] (0,1) rectangle (1,2);
    \fill[cardp1!12] (2,3) rectangle (4,4);
    \fill[cardp1!12] (3,2) rectangle (4,3);
    \draw[black!18, line width=0.25pt, step=1] (0,0) grid (4,4);
    \draw[cardp2!45, line width=0.45pt] (0,0) rectangle (2,1);
    \draw[cardp2!45, line width=0.45pt] (0,1) rectangle (1,2);
    \draw[cardp1!45, line width=0.45pt] (2,3) rectangle (4,4);
    \draw[cardp1!45, line width=0.45pt] (3,2) rectangle (4,3);
    \gcstoneone{0.5}{1.5}\gcstoneone{1.5}{0.5}\gcstoneone{2.5}{0.5}
    \gcstonetwo{1.5}{3.5}\gcstonetwo{3.5}{3.5}\gcstonetwo{3.5}{1.5}
    \draw[cardp1, line width=0.75pt, -latex] (0.5,1.5)--(0.5,2.5);
    \draw[cardp1, line width=0.75pt, -latex] (2.5,0.5)--(3.5,0.5);
    \draw[cardp2, line width=0.75pt, -latex] (3.5,1.5)--(3.5,0.5);
    \draw[cardp2, line width=0.75pt, -latex] (1.5,3.5)--(0.5,3.5);
  \end{tikzpicture}%
}

\newcommand{\gameboardKingsValley}{%
  \begin{tikzpicture}[scale=0.68]
    \gcgrid{4}
    \fill[cardp1!14] (1,1) rectangle (3,3);
    \draw[black!18, line width=0.25pt, step=1] (0,0) grid (4,4);
    \draw[cardink, line width=0.45pt] (0,0) rectangle (4,4);
    \gcchessking{1.5}{0.5}{cardp1}
    \gcchesspawn{2.5}{0.5}{cardp1}
    \gcchesspawn{3.5}{0.5}{cardp1}
    \gcchesspawn{3.5}{2.5}{cardp1}
    \gcchesspawn{0.5}{3.5}{cardp2}
    \gcchesspawn{1.5}{2.5}{cardp2}
    \gcchesspawn{2.5}{3.5}{cardp2}
    \gcchessking{3.5}{3.5}{cardp2}
    \draw[cardp1, line width=0.85pt, -latex] (1.5,0.5)--(1.5,1.5);
    \draw[cardp1!75, line width=0.65pt] (1.5,1.5) circle (0.31);
  \end{tikzpicture}%
}

\newcommand{\gameboardGroups}{%
  \begin{tikzpicture}[scale=0.36]
    \gcgrid{8}
    \foreach \x/\y in {1.5/6.5,2.5/4.5,3.5/4.5,4.5/4.5,4.5/3.5,5.5/3.5,5.5/2.5}{\gcstoneone{\x}{\y}}
    \foreach \x/\y in {3.5/5.5,5.5/5.5,2.5/3.5,3.5/3.5,6.5/3.5,4.5/2.5}{\gcstonetwo{\x}{\y}}
    \draw[cardp1, line width=1.0pt, rounded corners=2pt]
      (2.5,4.5)--(3.5,4.5)--(4.5,4.5)--(4.5,3.5)--(5.5,3.5)--(5.5,2.5);
  \end{tikzpicture}%
}

\newcommand{\gameboardTicTacChess}{%
  \begin{tikzpicture}[scale=0.78]
    \gcgrid{3}
    \foreach \x/\y/\c in {0.5/0.5/cardp1,1.5/1.5/cardp1,2.5/0.5/cardp1,0.5/2.5/cardp2,1.5/2.5/cardp2,2.5/1.5/cardp2}{\gcchessqueen{\x}{\y}{\c}}
    \draw[cardp1, line width=0.8pt, -latex] (0.5,0.5)--(2.5,2.5);
  \end{tikzpicture}%
}

\newcommand{\gameboardSimpleGame}{%
  \begin{tikzpicture}[scale=0.68]
    \gcgrid{4}
    \foreach \x/\y in {3.5/2.5,1.5/2.5,2.5/1.5,0.5/0.5}{\gcstoneone{\x}{\y}}
    \foreach \x/\y in {2.5/3.5,0.5/2.5,3.5/1.5,1.5/0.5}{\gcstonetwo{\x}{\y}}
    \draw[cardp1!45, line width=0.55pt, dashed] (2.5,0.5) circle (0.26);
    \draw[cardp2!45, line width=0.55pt, dashed] (0.5,3.5) circle (0.26);
    \draw[cardp1, line width=0.8pt, -latex] (2.5,0.5)--(2.5,1.5);
    \draw[cardp2, line width=0.8pt, -latex] (0.5,3.5)--(0.5,2.5);
  \end{tikzpicture}%
}

\newcommand{\gameboardAgapi}{%
  \begin{tikzpicture}[scale=0.68]
    \gcgrid{4}
    \foreach \x/\y in {0/3,0/1,1/1,2/0,3/0}{\gcblock{\x}{\y}}
    \foreach \x/\y/\c in {2.5/3.5/cardp1,3.5/1.5/cardp1,1.5/3.5/cardp2,1.5/0.5/cardp2}{%
      \node[font=\fontsize{15}{15}\selectfont,text=\c] at (\x,\y) {\symbishop};}
    \draw[cardp1, line width=0.8pt, -latex] (2.5,3.5)--(3.5,2.5);
    \draw[cardp1, line width=0.8pt, dashed, -latex] (3.5,2.5)--(3.5,0.5);
  \end{tikzpicture}%
}

\newcommand{\gameboardDomineering}{%
  \begin{tikzpicture}[scale=0.45]
    \gcgrid{6}
    \foreach \x/\y/\w/\h in {0.08/0.08/1.84/0.84,3.08/3.08/1.84/0.84,2.08/4.08/1.84/0.84}{%
      \filldraw[fill=cardp1!75, draw=cardink, line width=0.25pt] (\x,\y) rectangle ++(\w,\h);}
    \foreach \x/\y/\w/\h in {0.08/4.08/0.84/1.84,1.08/4.08/0.84/1.84,5.08/2.08/0.84/1.84,5.08/4.08/0.84/1.84}{%
      \filldraw[fill=cardp2!75, draw=cardink, line width=0.25pt] (\x,\y) rectangle ++(\w,\h);}
    \draw[cardink, line width=0.45pt] (0,0) rectangle (6,6);
  \end{tikzpicture}%
}

\newcommand{\gameboardFirstAttack}{%
  \begin{tikzpicture}[scale=0.45]
    \gcgrid{6}
    \gcstoneone{0.5}{0.5}
    \gcstonetwo{1.5}{2.5}
    \draw[cardp2, line width=0.65pt, dashed, opacity=0.75] (1.5,0)--(1.5,6);
    \draw[cardp2, line width=0.65pt, dashed, opacity=0.75] (0,2.5)--(6,2.5);
    \draw[cardp2, line width=0.55pt, dashed, opacity=0.45] (0,1)--(5,6);
    \draw[cardp2, line width=0.55pt, dashed, opacity=0.45] (0,4)--(4,0);
  \end{tikzpicture}%
}

\newcommand{\gameboardAlquerkonane}{%
  \begin{tikzpicture}[scale=0.68]
    \gcgrid{4}
    \foreach \x/\y in {0.5/0.5,2.5/0.5,1.5/1.5,3.5/1.5}{\gcstoneone{\x}{\y}}
    \foreach \x/\y in {0.5/3.5,2.5/3.5,1.5/2.5,3.5/2.5}{\gcstonetwo{\x}{\y}}
    \draw[cardp1, line width=0.75pt, -latex] (1.5,1.5)--(2.5,2.5);
    \draw[cardp1, line width=0.8pt, -latex] (3.70,1.5)--(3.70,3.5);
    \draw[cardp2!70, line width=0.7pt] (3.5,2.5) circle (0.30);
  \end{tikzpicture}%
}

\newcommand{\gameboardTicTacToe}{%
  \begin{tikzpicture}[scale=0.78]
    \gcgrid{3}
    \node[font=\Large\bfseries,text=cardp1] at (0.5,0.5) {X};
    \node[font=\Large\bfseries,text=cardp1] at (1.5,1.5) {X};
    \node[font=\Large\bfseries,text=cardp1] at (2.5,2.5) {X};
    \node[font=\Large\bfseries,text=cardp2] at (2.5,0.5) {O};
    \node[font=\Large\bfseries,text=cardp2] at (0.5,2.5) {O};
    \draw[cardp1, line width=1.0pt] (0.22,0.22)--(2.78,2.78);
  \end{tikzpicture}%
}

\newcommand{\gameboardConnectFour}{\connectfourboard[0.43]}

\newcommand{\gameboardOthello}{%
  \begin{tikzpicture}[scale=0.45]
    \gcgrid{6}
    \foreach \x/\y in {5.5/5.5,2.5/4.5,3.5/4.5,2.5/3.5,3.5/3.5,5.5/2.5,1.5/1.5,2.5/1.5,0.5/0.5,2.5/0.5}{\gcstoneone{\x}{\y}}
    \foreach \x/\y in {4.5/5.5,4.5/4.5,1.5/3.5,4.5/3.5,3.5/2.5,4.5/2.5,3.5/1.5,3.5/0.5}{\gcstonetwo{\x}{\y}}
    \draw[cardp1, line width=0.75pt, -latex] (1.5,3.5)--(2.5,3.5)--(3.5,3.5);
  \end{tikzpicture}%
}

\newcommand{\gameboardThreeSixNine}{%
  \begin{tikzpicture}[scale=0.30]
    \gcgrid{9}
    \foreach \x/\y in {0.5/0.5,1.5/1.5,2.5/2.5,3.5/3.5,4.5/4.5,5.5/5.5,6.5/6.5,7.5/7.5,8.5/8.5,6.5/1.5,6.5/2.5,6.5/3.5,6.5/5.5,6.5/7.5,0.5/4.5,2.5/4.5}{%
      \fill[cardp1] (\x,\y) circle (0.17);}
    \foreach \x/\y in {2.5/8.5,5.5/8.5,1.5/7.5,4.5/7.5,8.5/6.5,1.5/5.5,3.5/5.5,7.5/5.5,2.5/3.5,7.5/3.5,0.5/2.5,4.5/2.5,8.5/2.5,3.5/1.5,7.5/1.5}{%
      \fill[cardp2] (\x,\y) circle (0.17);}
    \draw[cardp2, line width=0.95pt] (0.5,0.5)--(8.5,8.5);
    \draw[cardp1, line width=0.95pt] (6.5,1.5)--(6.5,7.5);
    \draw[cardp1, line width=0.95pt] (0.5,4.5)--(4.5,4.5);
    \node[font=\fontsize{7}{7}\selectfont\bfseries,text=cardp2,fill=cardbeige,inner sep=1pt] at (8.45,8.75) {+3};
    \node[font=\fontsize{7}{7}\selectfont\bfseries,text=cardp1,fill=cardbeige,inner sep=1pt] at (6.5,8.05) {+2};
    \node[font=\fontsize{7}{7}\selectfont\bfseries,text=cardp1,fill=cardbeige,inner sep=1pt] at (4.85,4.5) {+1};
  \end{tikzpicture}%
}

\newcommand{\gameboardAralzaa}{%
  \begin{tikzpicture}[scale=0.78]
    \gcgrid{3}
    \foreach \x/\y in {0.5/0.5,2.5/0.5,1.5/2.5}{\gcchessknight{\x}{\y}{cardp1}}
    \foreach \x/\y in {0.5/2.5,2.5/2.5,1.5/0.5}{\gcchessknight{\x}{\y}{cardp2}}
    \draw[cardp1, line width=0.8pt, -latex, rounded corners=1.5pt] (2.5,0.5)--(2.5,1.5)--(0.5,1.5);
  \end{tikzpicture}%
}

\newcommand{\gameboardEpelle}{%
  \begin{tikzpicture}[scale=0.78]
    \draw[black!40, line width=0.75pt] (0,0) grid (2,2);
    \draw[black!40, line width=0.75pt] (0,0)--(2,2) (0,2)--(2,0);
    \foreach \x/\y in {0/0,1/0,2/0,0/1,1/1,2/1,0/2,1/2,2/2}{\gcemptypt{\x}{\y}}
    \gcstoneptone{2}{2}\gcstoneptone{1}{1}\gcstoneptone{2}{1}
    \gcstonepttwo{0}{1}\gcstonepttwo{0}{0}\gcstonepttwo{2}{0}
    \draw[cardp1, line width=0.95pt, rounded corners=1pt] (2,2)--(1,1)--(2,1);
    \draw[cardp2, line width=0.75pt, -latex] (2,0)--(1,0);
  \end{tikzpicture}%
}

\newcommand{\gameboardTapatan}{%
  \begin{tikzpicture}[scale=0.78]
    \draw[black!40, line width=0.75pt] (0,0) grid (2,2);
    \draw[black!40, line width=0.75pt] (0,0)--(2,2) (0,2)--(2,0);
    \foreach \x/\y in {0/0,1/0,2/0,0/1,1/1,2/1,0/2,1/2,2/2}{\gcemptypt{\x}{\y}}
    \gcstoneptone{0}{0}\gcstoneptone{1}{1}\gcstoneptone{2}{0}
    \gcstonepttwo{0}{2}\gcstonepttwo{1}{2}\gcstonepttwo{2}{1}
    \draw[cardp1, line width=1.0pt] (0,0)--(1,1)--(2,0);
  \end{tikzpicture}%
}

\newcommand{\gameboardChains}{%
  \begin{tikzpicture}[scale=0.88]
    \gchexcell{0}{0}{cardpaper}
    \gchexcell{0.63}{0.36}{cardpaper}
    \gchexcell{0.63}{-0.36}{cardp1!10}
    \gchexcell{0}{0.72}{cardp2!10}
    \gchexcell{0}{-0.72}{cardpaper}
    \gchexcell{-0.63}{0.36}{cardpaper}
    \gchexcell{-0.63}{-0.36}{cardpaper}
    \foreach \x/\y in {-0.63/0.36,0/0,0.63/0.36}{\fill[cardp1] (\x,\y) circle (0.15);}
    \foreach \x/\y in {0/0.72,0/-0.72}{\fill[cardp2] (\x,\y) circle (0.15);}
    \draw[cardp1, line width=1.0pt, rounded corners=1pt] (-0.63,0.36)--(0,0)--(0.63,0.36);
    \draw[cardp1, line width=0.8pt, -latex] (0,0)--(0.63,-0.36);
    \draw[cardp2!75, line width=0.65pt, dashed] (0,0.72) circle (0.23);
    \node[font=\fontsize{5}{5}\selectfont\bfseries,text=cardp1] at (0.08,0.17) {3};
    \node[font=\fontsize{5}{5}\selectfont\bfseries,text=cardp2] at (0.12,0.88) {1};
  \end{tikzpicture}%
}

\begin{gamecard}%
  {3 6 9}%
  {Shared-stone scoring game over lines of length three, six, and nine.}%
  {\gctagsThreeSixNine}
  \noindent
  \begin{minipage}[c]{\gamecardboardwidth}\raggedright\gameboardThreeSixNine\end{minipage}\hfill
  \begin{minipage}[c]{\gamecardruleswidth}
    \begin{gamerules}
      \gcsetup Players share an empty \(9\times9\) board; stones do not belong to either player after placement.
      \gcmove On each turn, place one shared stone on any empty cell.
      \gcrulekey The mover scores from non-contiguous occupied lines through the placed cell.
      \gcwin Lines of exactly three, six, or nine occupied cells score one, two, or three points; the higher final score wins.
    \end{gamerules}
    \par\smallskip
    \textit{Variation axes:} scoring thresholds and weights, orthogonal versus diagonal scoring, and center or corner locks.
  \end{minipage}
\end{gamecard}

\begin{gamecard}%
  {A Simple Game}%
  {Orthogonal movement game with a three-in-a-row objective.}%
  {\gctagsSimpleGame}
  \noindent
  \begin{minipage}[c]{\gamecardboardwidth}\raggedright\gameboardSimpleGame\end{minipage}\hfill
  \begin{minipage}[c]{\gamecardruleswidth}
    \begin{gamerules}
      \gcsetup Each player starts with four stones on alternating cells along the top and bottom edges.
      \gcmove Move one own stone exactly one orthogonal step into an adjacent empty cell.
      \gcrulekey Diagonal movement is not allowed, and there are no captures.
      \gcwin A player wins immediately after forming a horizontal, vertical, or diagonal line of three stones.
    \end{gamerules}
  \end{minipage}
\end{gamecard}

\begin{gamecard}%
  {Agapi}%
  {Bishop movement followed by a permanent orthogonal blocker shot.}%
  {\gctagsAgapi}
  \noindent
  \begin{minipage}[c]{\gamecardboardwidth}\raggedright\gameboardAgapi\end{minipage}\hfill
  \begin{minipage}[c]{\gamecardruleswidth}
    \begin{gamerules}
      \gcsetup Each player has two bishops on a \(4\times4\) board.
      \gcmove A turn first moves one bishop diagonally to the farthest reachable empty cell.
      \gcrulekey From the destination, the mover shoots a permanent blocker orthogonally to the farthest reachable empty cell.
      \gcwin If a player begins their move phase with no legal move-plus-shot option, the other player wins.
    \end{gamerules}
  \end{minipage}
\end{gamecard}

\begin{gamecard}%
  {All Queens Chess}%
  {Queen-slide race to form a line before the board locks.}%
  {\gctagsAllQueens}
  \noindent
  \begin{minipage}[c]{\gamecardboardwidth}\raggedright\gameboardAllQueens\end{minipage}\hfill
  \begin{minipage}[c]{\gamecardruleswidth}
    \begin{gamerules}
      \gcsetup Each player controls three queens on a \(4\times4\) board.
      \gcmove Choose a queen and slide it orthogonally through empty cells to the farthest reachable square.
      \gcrulekey The square the queen left becomes a permanent blocker that no queen may enter or cross.
      \gcwin A player wins by forming three queens in a row, column, or diagonal, or by leaving the opponent with no legal slide.
    \end{gamerules}
  \end{minipage}
\end{gamecard}

\begin{gamecard}%
  {Alquerkonane}%
  {Checkers-like forward movement with orthogonal jump captures.}%
  {\gctagsAlquerkonane}
  \noindent
  \begin{minipage}[c]{\gamecardboardwidth}\raggedright\gameboardAlquerkonane\end{minipage}\hfill
  \begin{minipage}[c]{\gamecardruleswidth}
    \begin{gamerules}
      \gcsetup The game starts from a fixed pawn pattern on a \(4\times4\) board.
      \gcmove A pawn may move one step diagonally forward to an empty square.
      \gcrulekey A pawn may instead jump orthogonally over an adjacent enemy pawn to the empty square beyond, removing it.
      \gcwin There is no promotion; a player wins by leaving the opponent with no legal move.
    \end{gamerules}
  \end{minipage}
\end{gamecard}

\begin{gamecard}%
  {Aralzaa}%
  {Forward-leap race to occupy the opponent's home row.}%
  {\gctagsAralzaa}
  \noindent
  \begin{minipage}[c]{\gamecardboardwidth}\raggedright\gameboardAralzaa\end{minipage}\hfill
  \begin{minipage}[c]{\gamecardruleswidth}
    \begin{gamerules}
      \gcsetup Each player starts with three horse-shaped pieces on their home row.
      \gcmove A piece leaps two squares forward and one square sideways into an empty square.
      \gcrulekey The forward knight-style leap does not capture; a player with no legal leap passes.
      \gcwin A player wins immediately after occupying all three squares of the opponent's starting row.
    \end{gamerules}
  \end{minipage}
\end{gamecard}

\begin{gamecard}%
  {Bajr}%
  {Directional camp-race game on a compact square board.}%
  {\gctagsBajr}
  \noindent
  \begin{minipage}[c]{\gamecardboardwidth}\raggedright\gameboardBajr\end{minipage}\hfill
  \begin{minipage}[c]{\gamecardruleswidth}
    \begin{gamerules}
      \gcsetup Player~1 starts in the lower-left camp and Player~2 starts in the upper-right camp.
      \gcmove Move one stone one orthogonal step into an empty cell.
      \gcrulekey Player~1 may move upward or sideways toward the right; Player~2 may move downward or sideways toward the left.
      \gcwin A player wins by occupying all three cells of the opponent's camp, or by leaving the opponent without a legal move.
    \end{gamerules}
  \end{minipage}
\end{gamecard}

\begin{gamecard}%
  {Chains of Thought}%
  {Hex-chain game where legal moves depend on local group sizes.}%
  {\gctagsChains}
  \noindent
  \begin{minipage}[c]{\gamecardboardwidth}\raggedright\gameboardChains\end{minipage}\hfill
  \begin{minipage}[c]{\gamecardruleswidth}
    \begin{gamerules}
      \gcsetup Players use a radius-one hex board with seven cells.
      \gcmove A turn either places a stone on a legal empty hex or moves one stone to an adjacent empty hex that merges friendly groups.
      \gcrulekey Placement and movement legality depend on the relative sizes of adjacent friendly and opposing chains.
      \gcwin After every move, smaller adjacent opposing chains can be captured; if no legal move remains, chain-size signatures decide the winner.
    \end{gamerules}
  \end{minipage}
\end{gamecard}

\begin{gamecard}%
  {Connect 4}%
  {Vertical connection game on a \(6\times7\) grid.}%
  {\gctagsConnectFour}
  \noindent
  \begin{minipage}[c]{\gamecardboardwidth}\raggedright\connectfourboard[0.41]\end{minipage}\hfill
  \begin{minipage}[c]{\gamecardruleswidth}
    \begin{gamerules}
      \gcsetup Players use a vertical \(6\times7\) grid and Player~1 moves first.
      \gcmove Choose one non-full column and drop a disc into it; gravity sends it to the lowest empty cell.
      \gcrulekey A move is illegal if the chosen column is already full.
      \gcwin A player wins immediately after connecting four own discs horizontally, vertically, or diagonally.
    \end{gamerules}
    \par\smallskip
    \textit{Variation axes:} win patterns including exact or gapped four, legal winning directions, opening restrictions, and locked columns.
  \end{minipage}
\end{gamecard}

\begin{gamecard}%
  {Domineering}%
  {Asymmetric domino-placement game of move exhaustion.}%
  {\gctagsDomineering}
  \noindent
  \begin{minipage}[c]{\gamecardboardwidth}\raggedright\gameboardDomineering\end{minipage}\hfill
  \begin{minipage}[c]{\gamecardruleswidth}
    \begin{gamerules}
      \gcsetup Players share an initially empty \(6\times6\) grid.
      \gcmove The mover places one domino on two adjacent empty squares.
      \gcrulekey One player may place only horizontal dominoes, while the other may place only vertical dominoes.
      \gcwin A placement may not overlap or extend off the board; the first player with no legal placement loses.
    \end{gamerules}
    \par\smallskip
    \textit{Variation axes:} player orientation assignments, shared orientation modes, contact and opening constraints, and blocked-state outcomes such as win, loss, or draw.
  \end{minipage}
\end{gamecard}

\begin{gamecard}%
  {Epelle}%
  {Adjacent-sliding alignment game where starting rows do not count.}%
  {\gctagsEpelle}
  \noindent
  \begin{minipage}[c]{\gamecardboardwidth}\raggedright\gameboardEpelle\end{minipage}\hfill
  \begin{minipage}[c]{\gamecardruleswidth}
    \begin{gamerules}
      \gcsetup Each player begins with three pieces already on opposite rows of a \(3\times3\) intersection graph.
      \gcmove Move one own piece to an adjacent connected empty intersection.
      \gcrulekey Starting rows do not count as wins; the line must be created by play.
      \gcwin A player wins by creating a line of three along the board lines.
    \end{gamerules}
  \end{minipage}
\end{gamecard}

\begin{gamecard}%
  {Feldja}%
  {Alignment-and-capture game on a three-ring point graph.}%
  {\gctagsFeldja}
  \noindent
  \begin{minipage}[c]{\gamecardboardwidth}\raggedright\gameboardFeldja\end{minipage}\hfill
  \begin{minipage}[c]{\gamecardruleswidth}
    \begin{gamerules}
      \gcsetup Each player starts with nine reserve stones on a 24-point three-ring graph.
      \gcmove Place reserve stones on empty points, then slide committed stones to adjacent empty points.
      \gcrulekey A vacated point closes and cannot be reused; completing a marked line removes one opposing stone.
      \gcwin A player wins by reducing the opponent to two or fewer total stones, or by leaving the opponent with no legal action.
    \end{gamerules}
  \end{minipage}
\end{gamecard}

\begin{gamecard}%
  {First Attack}%
  {Non-attacking placement puzzle played as a last-move contest.}%
  {\gctagsFirstAttack}
  \noindent
  \begin{minipage}[c]{\gamecardboardwidth}\raggedright\gameboardFirstAttack\end{minipage}\hfill
  \begin{minipage}[c]{\gamecardruleswidth}
    \begin{gamerules}
      \gcsetup Players share a \(6\times6\) board and all placed stones are neutral.
      \gcmove Place one shared stone on an empty cell.
      \gcrulekey A legal cell shares no row, column, or diagonal with any occupied cell.
      \gcwin The player who makes the last legal placement wins.
    \end{gamerules}
    \par\smallskip
    \textit{Variation axes:} forbidden-neighbor geometry, opening locks, and normal versus misere last-move outcomes, including orthogonal, diagonal, and knight offsets.
  \end{minipage}
\end{gamecard}

\begin{gamecard}%
  {Groups}%
  {Connection game where friendly jumps help assemble one group.}%
  {\gctagsGroups}
  \noindent
  \begin{minipage}[c]{\gamecardboardwidth}\raggedright\gameboardGroups\end{minipage}\hfill
  \begin{minipage}[c]{\gamecardruleswidth}
    \begin{gamerules}
      \gcsetup The game starts from a fixed \(8\times8\) position with seven Player~1 pieces and six Player~2 pieces.
      \gcmove A piece either steps orthogonally to an adjacent empty square or jumps one friendly piece to the empty square beyond.
      \gcrulekey Jumps do not capture; the jumped piece remains on the board.
      \gcwin A player wins immediately after at least six own pieces form a single orthogonally connected group.
    \end{gamerules}
  \end{minipage}
\end{gamecard}

\begin{gamecard}%
  {King's Valley}%
  {Sliding-piece race to bring the king into the central throne.}%
  {\gctagsKingsValley}
  \noindent
  \begin{minipage}[c]{\gamecardboardwidth}\raggedright\gameboardKingsValley\end{minipage}\hfill
  \begin{minipage}[c]{\gamecardruleswidth}
    \begin{gamerules}
      \gcsetup Each player starts with one king and three guards on a \(4\times4\) board.
      \gcmove Choose one piece and one orthogonal direction; the piece slides as far as possible through empty cells.
      \gcrulekey The four central cells form the throne zone.
      \gcwin A player wins when their king lands on a throne-zone cell, or by leaving the opponent with no legal slide.
    \end{gamerules}
  \end{minipage}
\end{gamecard}

\begin{gamecard}%
  {Moxie}%
  {Compact placement, movement, and forced-jump alignment game.}%
  {\gctagsMoxie}
  \noindent
  \begin{minipage}[c]{\gamecardboardwidth}\raggedright\gameboardMoxie\end{minipage}\hfill
  \begin{minipage}[c]{\gamecardruleswidth}
    \begin{gamerules}
      \gcsetup Each player starts with four reserve stones on an empty \(3\times3\) board.
      \gcmove If no capture is available, place a reserve stone or move one stone one orthogonal step.
      \gcrulekey If an orthogonal jump capture is available, the player must take one; captures do not chain.
      \gcwin A player wins by forming three in a row, reducing the opponent to two or fewer total stones, or immobilizing the opponent.
    \end{gamerules}
  \end{minipage}
\end{gamecard}

\begin{gamecard}%
  {Othello}%
  {Disc-flipping majority game with straight-line outflanks.}%
  {\gctagsOthello}
  \noindent
  \begin{minipage}[c]{\gamecardboardwidth}\raggedright\gameboardOthello\end{minipage}\hfill
  \begin{minipage}[c]{\gamecardruleswidth}
    \begin{gamerules}
      \gcsetup The game uses a \(6\times6\) board with the standard four-disc center start.
      \gcmove Place a disc on an empty square that outflanks one or more straight lines of opposing discs.
      \gcrulekey All outflanked discs immediately flip to the mover's color; a player with no legal placement passes.
      \gcwin When the board is full or both players are stuck, the higher disc count wins.
    \end{gamerules}
  \end{minipage}
\end{gamecard}

\begin{gamecard}%
  {Tapatan}%
  {Placement-then-sliding alignment game on a nine-point graph.}%
  {\gctagsTapatan}
  \noindent
  \begin{minipage}[c]{\gamecardboardwidth}\raggedright\gameboardTapatan\end{minipage}\hfill
  \begin{minipage}[c]{\gamecardruleswidth}
    \begin{gamerules}
      \gcsetup Each player has three stones on a \(3\times3\) point graph with diagonals.
      \gcmove First place stones on empty intersections, then slide one stone to an adjacent connected empty intersection.
      \gcrulekey Lines are checked along the board lines in both phases.
      \gcwin A player wins immediately after forming a line of three.
    \end{gamerules}
  \end{minipage}
\end{gamecard}

\begin{gamecard}%
  {Tic-Tac-Chess}%
  {Tic-tac-toe objective with queen movement after placement.}%
  {\gctagsTicTacChess}
  \noindent
  \begin{minipage}[c]{\gamecardboardwidth}\raggedright\gameboardTicTacChess\end{minipage}\hfill
  \begin{minipage}[c]{\gamecardruleswidth}
    \begin{gamerules}
      \gcsetup Players first alternate placing three queen pieces each on empty cells of a \(3\times3\) board.
      \gcmove After all six pieces are placed, a queen may slide any distance in a straight line through empty cells.
      \gcrulekey A queen may also hop over one adjacent enemy piece to the empty cell beyond; hops do not capture.
      \gcwin A player wins immediately after forming a horizontal, vertical, or diagonal line of three pieces.
    \end{gamerules}
    \par\smallskip
    \textit{Variation axes:} placement-phase timing, winning-line direction, piece movement, hop or capture behavior, and center or corner locks.
  \end{minipage}
\end{gamecard}

\begin{gamecard}%
  {Tic-Tac-Toe}%
  {Classic mark-placement alignment game on a \(3\times3\) grid.}%
  {\gctagsTicTacToe}
  \noindent
  \begin{minipage}[c]{\gamecardboardwidth}\raggedright\gameboardTicTacToe\end{minipage}\hfill
  \begin{minipage}[c]{\gamecardruleswidth}
    \begin{gamerules}
      \gcsetup Player~1 uses X and moves first; Player~2 uses O.
      \gcmove Place one mark on any empty cell.
      \gcrulekey A move is legal only if the chosen cell is empty.
      \gcwin A player wins by forming a line of three; a full board without a line is a draw.
    \end{gamerules}
  \end{minipage}
\end{gamecard}

\section{Corpus Statistics}
\label{app:stats}

\begin{figure}[ht]
  \centering
  \resizebox{.98\linewidth}{!}{%
  \begin{tikzpicture}
    \begin{axis}[
      xbar,
      width=.92\linewidth,
      height=4.2cm,
      xmin=0,
      xmax=16,
      xlabel={variant count},
      symbolic y coords={opening or locked-cell edits,scoring or objective edits,movement or attack-rule edits,directionality constraints},
      ytick=data,
      y dir=reverse,
      enlarge y limits=0.18,
      major grid style={dashed,gray!45},
      xmajorgrids=true,
      axis background/.style={fill=rqplotbg},
      every tick label/.append style={font=\scriptsize},
      yticklabel style={align=right,text width=2.65cm},
      xlabel style={font=\scriptsize},
      nodes near coords,
      every node near coord/.append style={font=\scriptsize,anchor=west,xshift=2pt},
      bar width=8pt
    ]
      \addplot+[fill=qwenThreeFiveLine!76,draw=qwenThreeFiveLine!90!black] coordinates {
        (14,opening or locked-cell edits)
        (11,scoring or objective edits)
        (6,movement or attack-rule edits)
        (14,directionality constraints)
      };
    \end{axis}
  \end{tikzpicture}}
  \caption{\textbf{Controlled variant distribution.} The 45 controlled variants grouped by edited rule family.}
  \label{fig:app-variant-families}
\end{figure}
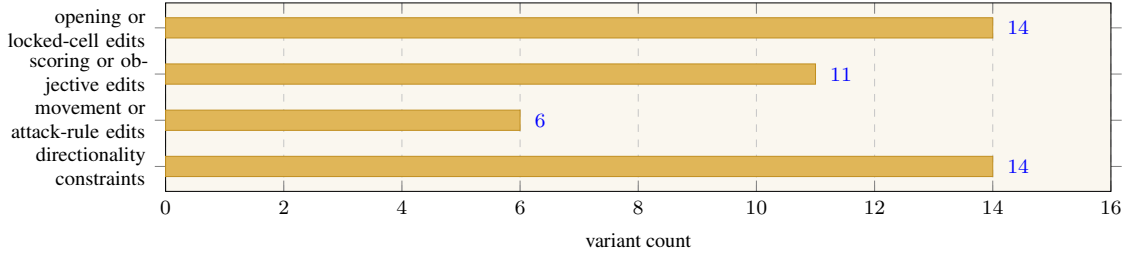

The controlled-diversity pool is separate from the four-game corpus that supplies the main \SPSD conditions, and its 15 held-out games never contribute training rows.
Figure~\ref{fig:app-variant-families} groups the 45 variants by the rule family they edit, and Table~\ref{tab:app-corpus-stats} reports the structural range of the 50 environments.

\begin{table}[ht]
  \centering
  \scriptsize
  \setlength{\tabcolsep}{4pt}
  \begin{tabularx}{\linewidth}{@{}lXXXX@{}}
    \toprule
    \textbf{Statistic} & \textbf{Min} & \textbf{Median} & \textbf{Mean} & \textbf{Max} \\
    \midrule
    Max game length & 12 & 42 & 50.2 & 81 \\
    Action-space size & 7 & 36 & 47.0 & 81 \\
    Board cells & 9 & 36 & 40.8 & 81 \\
    Board rows & 3 & 6 & 6.0 & 9 \\
    Board cols & 3 & 6 & 6.2 & 9 \\
    \bottomrule
  \end{tabularx}
  \caption{\textbf{Diversity-pool statistics.} Min, median, mean, and max for the structural properties of the 50 environments.}
  \label{tab:app-corpus-stats}
\end{table}

\section{Training Hyperparameters}
\label{app:hyperparameters}

\Cref{tab:app-hparams} reports the settings used to construct controlled game variants, train the ExIt planning experts, build search traces, and post-train the Qwen3-4B \SPSD student.
The controlled-diversity procedure constructs candidate games and variants for the held-out transfer study, whereas the four EfficientZero experts used to generate the main \SPSD corpus are trained under the settings in \Cref{app:config,app:muzero-training}.
We keep trace-construction parameters fixed across ablations so that differences come from the source game, downstream training objective, or teacher MCTS budget.
\begin{table}[ht]
  \centering
  \scriptsize
  \setlength{\tabcolsep}{4pt}
  \begin{tabularx}{\linewidth}{@{}lXX@{}}
    \toprule
    \textbf{Component} & \textbf{Hyperparameter} & \textbf{Value} \\
    \midrule
    Variant authoring & Base games & Connect4, Tic-Tac-Chess, 3 6 9, Domineering, First Attack \\
    Variant authoring & Variants per base game & 9 \\
    Held-out evaluation & Evaluation games & 15 held-out games \\
    ExIt-trained planning expert & MCTS simulations per move & 32, 64, 128, 256 \\
    ExIt-trained planning expert & Exploration constant & $c_{\mathrm{puct}}=1.5$ \\
    Trace construction & Max retained root alternatives & $K=3$ \\
    Trace construction & Max verbalized search depth & $D=5$ \\
    Student \SPSD & Base model & Qwen3-4B \\
    Student \SPSD & Sequence length & 4096 tokens \\
    Student \SPSD & Learning rate & $2 \times 10^{-5}$ \\
    Student \SPSD & Epochs & 2 \\
    Student \SPSD & Effective batch size & 128 examples \\
    \bottomrule
  \end{tabularx}
  \caption{\textbf{Training configuration.} The table reports hyperparameters for variant authoring, ExIt planning experts, trace construction, and Qwen3-4B \SPSD post-training.}
  \label{tab:app-hparams}
\end{table}

\cref{tab:opsd-search-space} records every scientific and optimization choice explored for \OPSD.
The asterisks identify the configuration used for the main matched experiments, and the daggers identify the configuration that reached the highest mathematics transfer in the search.
Hardware topology and inference-server memory allocation are runtime settings and therefore remain outside the selection grid.
\begin{table}[!htb]
  \centering
  \scriptsize
  \setlength{\tabcolsep}{4pt}
    \begin{tabularx}{\linewidth}{@{}lX@{}}
      \toprule
      \textbf{Hyperparameter} & \textbf{Search space} \\
      \midrule
      Student model & \{Qwen3-4B-Base$\ast\dag$, Qwen3-8B$\ast$, Llama-3.1-8B-Instruct$\ast$, Qwen3-30B-A3B-Instruct\} \\
      Learning rate & \{$2\times10^{-6}$, $3\times10^{-6}\dag$, $5\times10^{-6}\ast$, $1\times10^{-5}$\} \\
      Learning-rate schedule & \{constant$\ast\dag$, cosine with 0.03 warmup\} \\
      Generalized-JSD $\beta$ & \{0, 0.10, $0.25\ast\dag$, 0.5, 1.0\} \\
      Teacher update & \{fixed$\ast$, EMA$\dag$\} \\
      EMA teacher decay & \{0.995, 0.996, 0.997, $0.998\dag$, 0.999\} \\
      Distillation estimator & \{full-vocabulary$\ast\dag$, sampled-token reverse KL\} \\
      Optimizer steps & \{400, $1{,}000\ast\dag$, 2,000\} \\
      Effective batch size & \{8, $16\ast\dag$, 32\} \\
      LoRA adapter & \{$r{=}64,\alpha{=}128\ast$, $r{=}128,\alpha{=}256\dag$, $r{=}256,\alpha{=}512$\}, dropout 0 \\
      Maximum completion length & 1,024 tokens \\
      Maximum sequence length & 20,000 tokens \\
      Gradient norm clip & 0.1 \\
      Distillation weight $\lambda$ & 1.0 \\
      Per-token divergence clip & 0.06 \\
      Rollout sampling & $T=1.1$, $p=0.95$, $k=20$ \\
      Student / teacher thinking & disabled / enabled \\
      Random seed & 42 \\
      \bottomrule
    \end{tabularx}
  \caption{\textbf{On-policy self-distillation hyperparameter search.} The table reports every explored training choice and the settings held fixed. $\ast$ marks the matched main-experiment configuration and $\dag$ the configuration with the highest mathematics transfer in \Cref{tab:opsd-teacher-search}.}
  \label{tab:opsd-search-space}
\end{table}

\subsection{Teacher and Optimization Selection}
\label{app:opsd-selection}

\Cref{tab:opsd-teacher-search} reports the outcome of the search on Qwen3-4B-Base, the condition with the widest training dynamic range.
Every run uses the same corpus, prompt contract, decoding, seed, and 1,000-step budget.
These runs use a single evaluation configuration that differs from the one behind \Cref{tab:main-results}, so their absolute values are not directly comparable to the main table.
\begin{table}[!htb]
  \centering
  \scriptsize
  \setlength{\tabcolsep}{4pt}
  \begin{tabularx}{\linewidth}{@{}lXr@{}}
    \toprule
    \textbf{Teacher} & \textbf{Varied factor} & \textbf{Final} \\
    \midrule
    Fixed & $\beta=0.25$, lr $5\times10^{-6}$, $r{=}64$ (main experiments) & 31.9 \\
    Fixed & Cosine schedule & 31.5 \\
    Fixed & Forward KL ($\beta=0$) & 4.0 \\
    Fixed & lr $2\times10^{-6}$ & 34.5 \\
    \midrule
    EMA 0.999 & lr $5\times10^{-6}$, $r{=}64$ & 35.0 \\
    EMA 0.999 & lr $3\times10^{-6}$ & 34.2 \\
    EMA 0.999 & lr $2\times10^{-6}$ & 32.5 \\
    EMA 0.995 & lr $5\times10^{-6}$ & 35.0 \\
    EMA 0.999 & lr $3\times10^{-6}$, $r{=}128$ & 35.6 \\
    EMA 0.999 & $r{=}256$ & 35.5 \\
    EMA 0.999 & Effective batch 32 & 34.7 \\
    EMA 0.999 & $\beta=0.10$ & 34.5 \\
    EMA 0.996 & $r{=}128$ & 35.7 \\
    EMA 0.997 & $r{=}128$ & 36.2 \\
    EMA 0.997 & $r{=}256$ & 35.4 \\
    \textbf{EMA 0.998}$\dag$ & $r{=}128$ & \textbf{36.6} \\
    EMA 0.997 & $r{=}128$, 2,000 steps & 33.2 \\
    \bottomrule
  \end{tabularx}
  \caption{\textbf{Teacher and optimization selection on Qwen3-4B-Base.} Mean accuracy over the six mathematics benchmarks at the end of training, for runs that differ from the row above the midrule only in the factor named in the \textbf{Varied factor} column. All EMA rows use $\beta=0.25$ and learning rate $3\times10^{-6}$ unless stated.}
  \label{tab:opsd-teacher-search}
\end{table}
An exponential moving-average teacher outperforms the fixed teacher used in the main experiments, and its decay follows an inverted U with an optimum at 0.998, which reaches 36.6 against 31.9 for the matched fixed-teacher run.
Adapter capacity beyond $r{=}128$, larger batches, a weaker divergence anchor, and doubling the step budget all leave that level unreached, and forward KL alone collapses training.

\section{Prompt Templates}
\label{app:prompts}

All training conversations follow the same contract.
The system message contains the rules, the user message contains the board and the legal actions, and the assistant must end with a boxed legal move.
In the replay analysis, that final boxed move is the selected action evaluated by the fresh 50-simulation MuZero MCTS search described in \cref{sec:metrics-analysis}.
Figure~\ref{fig:game-prompt-template} shows the field order of the move prompt and the \SPSD ASCII board format shared by every game.
\paragraph{Move-prompt template.}
Each game supplies its own rules, state lines, board text, option handles, and option descriptions within this fixed field order. Line breaks in the figure are typographic.
\begin{figure}[ht]
  \centering
  \small
  \begin{tcolorbox}[colframe=black, colback=gray!5, title=Game-Playing Prompt Template]
    {\ttfamily\footnotesize
    Game Rules:\\
    \{rules\_text\}\\[4pt]

    Player to move: Player \{current\_player\} (\{X\_or\_O\}).\\[4pt]

    Current State:\\
    \{state\_lines\}\\[4pt]

    Legend:\\
    \{legend\_lines\}\\[4pt]

    Current Board:\\
    \{ascii\_board\}\\[4pt]

    Legal Options:\\
    - \{handle\}: \{human\_description\}\\[4pt]

    Your objective is to choose the legal move that gives you the
    best game result: win if possible, otherwise draw rather than lose.\\
    A legal handle is the exact text after `- ' and before `:' in one
    Legal Options bullet, or the full bullet text after `- ' when that
    bullet has no `:'. Copy the legal handle verbatim, including leading
    words such as `place' or `column'; do not abbreviate it to only a
    coordinate or number.\\
    Reason briefly, then choose exactly one legal handle from Legal
    Options. End with that legal handle inside
    \textbackslash boxed\{\}.}
  \end{tcolorbox}
  \caption{\textbf{Game-playing prompt template.} The prompt provides the optional game rules, current state and board representation, and the available legal actions. The model is instructed to reason briefly and return exactly one legal action using its full handle inside a boxed answer.}
  \label{fig:game-prompt-template}
\end{figure}
The non-reasoning contract keeps the same visible fields and replaces the final two lines with ``Choose exactly one legal handle from Legal Options. Return only that legal handle inside \(\boxed{}\).''
For state questions we append a task-specific question instead of the move-selection instruction, and Table~\ref{tab:app-spsd-dataset-taxonomy} lists the six question forms and answer types.
Figure~\ref{fig:connect4-prompt} instantiates the same contract for a single Connect4 position.
\begin{figure}[ht]
  \centering
  \small
  \begin{tcolorbox}[colframe=black, colback=gray!5, title=Connect4 \SPSD Training Example]
    {\ttfamily\footnotesize
    System: Connect4 is a two-player connection game on a vertical 6x7 grid.
 Players choose a non-full column, the disc falls to the lowest empty cell, and the first player to connect four horizontally, vertically, or diagonally wins.\\[4pt]
    User: Let's play Connect4.
 I'll be Player 2 and you be Player 1.\\
    Description: Gravity matters, central columns support more future lines, and the final answer must be a legal boxed move.\\
    Board:\\
    \hspace*{1.2em}6 . . . . . . .\\
    \hspace*{1.2em}5 . . . . . . .\\
    \hspace*{1.2em}4 . . . . . . .\\
    \hspace*{1.2em}3 X O X . . . .\\
    \hspace*{1.2em}2 O X O X . . .\\
    \hspace*{1.2em}1 X O X O X . .\\
    \hspace*{1.2em}\ \ a b c d e f g\\
    Side to move: Player 1.\\
    Legal actions: column a, column b, column c, column d, column e, column f, column g.\\
    Choose the expert move and finish with \textbackslash boxed\{action\}.\\[4pt]
    Assistant: Column d is the cleanest continuation because the searched alternatives leave weaker replies: column c lets Player 2 answer in d and flatten the central threat, while column e keeps material but does not create an immediate fork. After d, the retained rollout still leaves Player 1 with a horizontal threat on row 1 and a diagonal follow-up if Player 2 blocks only one line.
 Therefore the move is \textbackslash boxed\{column d\}.}
  \end{tcolorbox}
  \caption{\textbf{Representative Connect4 prompt.} The example shows full rules in the system message, the board and legal actions in the user turn, and a move explanation ending with its boxed legal handle.}
  \label{fig:connect4-prompt}
\end{figure}

\section{RuleBot-Distill Example}
\label{app:rulebot-distill-example}

RuleBot-Distill uses a rule-bot-verified target to provide a compact reference completion while keeping the student prompt target-hidden.
\Cref{fig:rulebot-distill-example} separates the student-visible board task from the teacher-only rule-bot context and the resulting reference completion.

\begin{figure}[ht]
  \centering
  \begin{tcolorbox}[colframe=black,colback=gray!5,title=RuleBot-Distill Connect4 Example]
    \ttfamily\footnotesize
    \textbf{Student-visible prompt}
    \par
    Game: Connect4 on a vertical 6x7 grid. Player 2 (O) moves; a disc falls to the lowest empty cell in its chosen column.
    \par
    Current board, rows 6 through 1:
    \par
    6 . . . X . . .\quad 5 . . . O . . .\quad 4 . . O X . . X
    \par
    3 . . O O . X X\quad 2 . . O O . X X\quad 1 . . X X . O O
    \par
    Legal actions: column 1, column 2, column 3, column 5, column 6, column 7.
    \par
    Choose exactly one legal handle and end with the boxed answer.
    \par\medskip
    \textbf{Teacher-only RuleBot-Distill context}
    \par
    Verified action column 3 resolves to c5 while completing the unique vertical line of at least 4.
    \par\medskip
    \textbf{Reference completion}
    \par
    The verified action column 3 resolves to c5 while completing the unique vertical line of at least 4.
    \par
    Final answer: \(\boxed{\textnormal{column 3}}\)
  \end{tcolorbox}
  \caption{\textbf{RuleBot-Distill record.} The student receives rules, state, legal handles, and the answer contract but not the selected target. Teacher-only context verifies the target consequence, and the reference completion returns its boxed handle.}
  \label{fig:rulebot-distill-example}
\end{figure}

The prompt alone is supplied at inference. The privileged context builds and verifies the reference completion during teacher-side construction and never enters the student input. This arm uses rule-bot supervision without MCTS search.

\section{Grounded MCTS Example}
\label{app:grounded-mcts-example}

\Cref{fig:grounded-mcts-example} shows how a grounded record is presented. The student receives the board task in the visible prompt, while the grounded comparison and replay evidence remain in teacher-only context.
The displayed teacher context uses the vocabulary of the trace without exposing raw search telemetry.

\begin{figure}[ht]
  \centering
  \begin{tcolorbox}[colframe=black,colback=gray!5,title=Grounded MCTS Connect4 Example]
    \ttfamily\footnotesize
    \textbf{Student-visible prompt}
    \par
    Game: Connect4 on a vertical 6x7 grid. Player 2 (O) moves; a disc falls to the lowest empty cell in its chosen column.
    \par
    Current board, rows 6 through 1:
    \par
    6 . . . . . . .
    \par
    5 . . . . . . .
    \par
    4 . . . . . . .
    \par
    3 X . . . . . .
    \par
    2 O . . . . . .
    \par
    1 X X O O X . .
    \par
    Legal actions: column 1, column 2, column 3, column 4, column 5, column 6, column 7.
    \par
    Choose exactly one legal handle and end with the boxed answer.
    \par\medskip
    \textbf{Teacher-only grounded context}
    \par
    The fresh root comparison retains target column 4 and the most tempting legal alternative, column 7. A bounded replay verifies opponent column 1 as a legal, non-terminal reply after the target, after which the turn returns to Player 2.
    \par\medskip
    \textbf{Reference completion}
    \par
    Column 4 is preferred to the tempting column 7. The verified opponent reply in column 1 remains non-terminal and returns the turn to Player 2.
    \par
    Final answer: \(\boxed{\textnormal{column 4}}\)
  \end{tcolorbox}
  \caption{\textbf{Grounded MCTS record.} The record shows the student receiving rules, state, legal handles, and the answer contract, while teacher-only context contrasts the target with an alternative through bounded replay.
  The reference completion ends with the boxed handle, and numerical search telemetry remains hidden.}
  \label{fig:grounded-mcts-example}
\end{figure}

Only the prompt is used at inference. The grounded context is available during teacher-side construction and checking, while the student must infer the target from the visible state and legal-action contract.

\section{SPSD-dataset Task Taxonomy}
\label{app:spsd-dataset-taxonomy}

The six state-question tasks use the same replay-backed answer contract while probing different state facts.

\begin{table}[ht]
  \centering
  \scriptsize
  \setlength{\tabcolsep}{3pt}
  \begin{tabularx}{\linewidth}{@{}lXXX@{}}
    \toprule
    \textbf{Task} & \textbf{Question} & \textbf{Verifier-backed evidence} & \textbf{Expected answer} \\
    \midrule
    Occupancy & What occupies a named cell? & Replay reads the encoded occupant at the requested coordinate. & Empty, or the encoded player value. \\
    Legality & Is a named handle legal in this state? & Membership in the replay-enumerated legal-action list. & Yes or no. \\
    Threat count & How many immediate winning actions does the opponent have? & Opponent replies are enumerated and checked for immediate terminal wins. & A nonnegative integer. \\
    Legal-action count & How many legal actions does the current player have? & The replay-enumerated legal-action list is counted. & An integer. \\
    Legal-action enumeration & List every exact legal handle in displayed order. & The replay-verified legal handles are matched to the displayed list. & A comma-separated list of exact handles. \\
    Successor state & After a named action, what occupies a named successor cell? & The action is replayed and the requested successor cell is inspected. & Empty, or the encoded player value. \\
    \bottomrule
  \end{tabularx}
  \caption{\textbf{State-task taxonomy.} The table compares six \SPSD tasks by question type, replay-computed evidence, and compact expected answer.}
  \label{tab:app-spsd-dataset-taxonomy}
\end{table}

\end{document}